# *Abstract Answer Set Solvers with Backjumping and Learning (long version)*

YULIYA LIERLER

*Department of Computer Science*
*University of Texas at Austin*
*1 University Station C0500*
*Taylor Hall 2.124*
*Austin, USA*
*E-mail: yuliya@cs.utexas.edu*

## Abstract

Nieuwenhuis, Oliveras, and Tinelli (2006) showed how to describe enhancements of the Davis-Putnam-Logemann-Loveland algorithm using transition systems, instead of pseudocode. We design a similar framework for several algorithms that generate answer sets for logic programs: SMODELS, SMODELS$_{cc}$, ASP-SAT with Learning (CMODELS), and a newly designed and implemented algorithm SUP. This approach to describing answer set solvers makes it easier to prove their correctness, to compare them, and to design new systems.

## 1 Introduction

Answer Set Programming (ASP) is a methodology commonly used for solving combinatorial search problems (Lifschitz 2008). In the development of ASP solvers, computational ideas behind SAT solvers (Gomes et al. 2008) play an important role. Influence of SAT solvers development on ASP systems is twofold. On the one hand, such ASP solvers as ASSAT[1] and CMODELS[2] follow the so called SAT-based approach where a SAT solver is invoked for search, possibly multiple times. On the other hand, "native" ASP solvers that implement search procedures specifically suited for logic programs often adopt computational techniques from SAT solvers. For instance, DLV[3] implements backjumping (Ricca et al. 2006), and SMODELS$_{cc}$[4] (Ward and Schlipf 2004) extends the answer set solver SMODELS[5] by introducing restarts, conflict-driven backjumping, learning, and forgetting – techniques widely used in SAT solvers. The ASP solver SUP[6] (Lierler 2008) implements these features also.

---

[1] http://assat.cs.ust.hk/ .
[2] http://www.cs.utexas.edu/users/tag/cmodels .
[3] http://www.dbai.tuwien.ac.at/proj/dlv/ .
[4] http://www.nku.edu/∼wardj1/Research/smodels_cc.html .
[5] http://www.tcs.hut.fi/Software/smodels/ .

In this paper our main goal is to show how the "abstract" approach to describing SAT solvers proposed in (Nieuwenhuis et al. 2006) can be extended to ASP solvers that use these sophisticated features. Usually computation procedures are described in terms of pseudocode. In (Nieuwenhuis et al. 2006), the authors proposed an alternative approach to describing DPLL-like procedures. They introduced an abstract framework that captures what "states of computation" are, and what transitions between states are allowed. In this way, it defines a directed graph such that every execution of the DPLL procedure corresponds to a path in this graph. Some edges may correspond to unit propagation steps, some to branching, some to backtracking. This allows the authors to model a DPLL-like algorithm by a mathematically simple and elegant object, graph, rather than a collection of pseudocode statements. In (Lierler 2008), we extended this framework for describing such ASP algorithms as SMODELS, ASP-SAT with Backtracking, and SUP *without Learning*. In this paper, we expand our previous work on abstract answer set solvers to cover such features as backjumping and learning (and also forgetting and restart). We start by introducing an abstract framework that captures a general mechanism of these sophisticated features in ASP solvers. For instance, this framework provides the transition underlying the process of learning a clause, but it does not suggest which clause shall be learned. Similarly, it provides a general description of backjumping but it does not supply the means for computing a "backjump clause" necessary for an answer set solver to perform backjumping. We then enhance this abstract framework to capture enough information about a state of computation for deriving a backjump clause.

Usually, DPLL-like procedures implement conflict-driven backjumping and learning where a particular learning schema such as, for instance, *Decision* or *FirstUIP* (Mitchell 2005) is applied for computing a special kind of a backjump clause. There are two common methods for describing a backjump clause construction. One employs the implication graph (Marques-Silva and Sakallah 1996) and the other employs resolution (Mitchell 2005). Ward and Schlipf (2004) extended the notion of an implication graph to the SMODELS algorithm. They then defined an algorithm for computing *FirstUIP* backjump clauses utilized by SMODELS$_{cc}$ to implement conflict-driven backjumping and learning. In this paper we introduce the algorithms *BackjumpClause* and *BackjumpClauseFirstUIP* based on resolution and the enhanced abstract framework that compute *Decision* and *FirstUIP*[7] backjump clauses respectively.

In (Lierler 2008), we introduced the basic algorithm underlining the system SUP but neglected some of its features: conflict-driven backjumping, learning, forgetting, and restarts. Here we account for these techniques and use an abstract framework designed in this paper for describing system SUP. We emphasize that the work on this abstract framework helped us to develop ASP solver SUP, to incorporate learning into its algorithm, and to prove its correctness. We analyzed performance of SUP against such answer set solvers as CMODELS, SMODELS, SMODELS$_{cc}$, and CLASP[8]. Overall, SUP performs well against these rival systems.

---

[6] http://www.cs.utexas.edu/users/tag/sup .
[7] The names of the backjump clauses follow (Mitchell 2005).

We start the paper with Section 2 that reviews the abstract DPLL framework introduced in (Nieuwenhuis et al. 2006) and some logic programming concepts. In Section 3, we define a graph representing the application of the algorithm for finding supporting models of a logic program. This paves the way to defining a graph representing the application of the SMODELS algorithm to a program in Section 4. Section 4.2 elaborates on the relationship between previously defined abstract frameworks. Section 5 extends the abstract DPLL framework by introducing an additional inference rule so that the *generate and test* algorithm of the SAT-based ASP system CMODELS may be characterized by this graph. In Section 6, we review the abstract framework that describes DPLL enhanced by backjumping and learning. In Section 7, we define a general abstract framework for describing ASP algorithms that implement such phenomena as backjumping and learning. In Section 7.2 we describe the algorithms of systems SMODELS$_{cc}$ and SUP by means of this framework. In Section 8 we extend the abstract *generate and test* framework to accommodate backjumping and learning, and in Section 8.2 we use these findings to describe the CMODELS algorithm. Section 9 extends the framework to capture additional information about a computation state of a solver, states the correctness results, and describes how the frameworks are related to each other. Section 10 provides the proofs for these results. In Section 10.3 and 11 we introduce the algorithms based on the extended framework for computing a backjump clause that are important in implementing conflict-driven backjumping and learning. In Section 12 we introduce the concept of an extended graph for the *generate and test* abstract framework and state the correctness results. Section 13 provides the proofs for these results. At last, in Section 14 we provide the experimental analysis that compares performance of SUP with other answer set solvers.

## 2 Review: Abstract DPLL and Logic Programs

### 2.1 Abstract Classical DPLL

For a set $\sigma$ of atoms, a *record* $M$ relative to $\sigma$ is a list of literals over $\sigma$ where

(i) some literals in $M$ are annotated by $\Delta$ that marks them as *decision* literals,
(ii) $M$ contains no repetitions.

The concatenation of two such lists is denoted by juxtaposition. Frequently, we consider a record as a set of literals, ignoring both the annotations and the order between its elements. A literal $l$ is *unassigned by* a record if neither $l$ nor its complement $\overline{l}$ belongs to it.

---

[8] http://www.cs.uni-potsdam.de/clasp/ .

*Unit Propagate*:
$$M \Longrightarrow M\, l \quad \text{if} \quad \begin{cases} C \vee l \in F \text{ and} \\ \overline{C} \subseteq M \end{cases}$$

*Decide*:
$$M \Longrightarrow M\, l^\Delta \quad \text{if} \quad \begin{cases} M \text{ is consistent and} \\ l \text{ is unassigned by } M \end{cases}$$

*Fail*:
$$M \Longrightarrow \textit{FailState} \quad \text{if} \quad \begin{cases} M \text{ is inconsistent and} \\ M \text{ contains no decision literals} \end{cases}$$

*Backtrack*:
$$P\, l^\Delta\, Q \Longrightarrow P\, \overline{l} \quad \text{if} \quad \begin{cases} P\, l^\Delta\, Q \text{ is inconsistent, and} \\ Q \text{ contains no decision literals} \end{cases}$$

Fig. 1. The transition rules of the graph $\text{DP}_F$.

A *state* relative to $\sigma$ is either a distinguished state *FailState* or a record relative to $\sigma$. For instance, the states relative to a singleton set $\{a\}$ of atoms are

$$\textit{FailState},\ \emptyset,\ a,\ \neg a,\ a^\Delta,\ \neg a^\Delta, a\neg a,\ a^\Delta \neg a,$$
$$a\neg a^\Delta,\ a^\Delta \neg a^\Delta, \neg a\, a,\ \neg a^\Delta\, a,\ \neg a\, a^\Delta,\ \neg a^\Delta\, a^\Delta,$$

where by $\emptyset$ we denote the empty list.

If $C$ is a disjunction (conjunction) of literals then by $\overline{C}$ we understand the conjunction (disjunction) of the complements of the literals occurring in $C$. We will sometimes identify $C$ with the multi-set of its elements.

For any CNF formula $F$ (a finite set of clauses), we will define its *DPLL graph* $\text{DP}_F$. The nodes of $\text{DP}_F$ are the states relative to the set of atoms occurring in $F$. We use the terms "state" and "node" interchangeably. Recall that a node is called *terminal* in a graph if there is no edge leaving this node in the graph. If a state is consistent and complete then it represents a truth assignment for $F$.

The set of edges of $\text{DP}_F$ is described by a set of "transition rules." Each transition rule is an expression $M \Longrightarrow M'$ followed by a condition, where $M$ and $M'$ are nodes of $\text{DP}_F$. Whenever the condition is satisfied, the graph contains an edge from node $M$ to $M'$. Generally, an edge in the graph may be justified by several transition rules. Figure 1 presents four transition rules that characterize the edges of $\text{DP}_F$.

This graph can be used for deciding the satisfiability of a formula $F$ simply by constructing an arbitrary path leading from node $\emptyset$ until a terminal node $M$ is reached. The following proposition shows that this process always terminates, that $F$ is unsatisfiable if $M$ is *FailState*, and that $M$ is a model of $F$ otherwise.

*Proposition 1*
For any CNF formula $F$,

(a) graph $\text{DP}_F$ is finite and acyclic,

(b) any terminal state of $\text{DP}_F$ other than *FailState* is a model of $F$,
(c) *FailState* is reachable from $\emptyset$ in $\text{DP}_F$ if and only if $F$ is unsatisfiable.

For instance, let $F$ be the set consisting of the clauses

$$a \vee b$$
$$\neg a \vee c.$$

Here is a path in $\text{DP}_F$:

$$\begin{array}{ll}
\emptyset & \Longrightarrow (Decide) \\
a^\Delta & \Longrightarrow (Unit\ Propagate) \\
a^\Delta\, c & \Longrightarrow (Decide) \\
a^\Delta\, c\, b^\Delta &
\end{array} \quad (1)$$

The name of the transition rule after each $\Longrightarrow$ shows which rule justifies the presence of this edge in the graph. Since the state $a^\Delta\, c\, b^\Delta$ is terminal, Proposition 1(b) asserts that $\{a, c, b\}$ is a model of $F$. Here is another path in $\text{DP}_F$ from $\emptyset$ to the same terminal node:

$$\begin{array}{ll}
\emptyset & \Longrightarrow (Decide) \\
a^\Delta & \Longrightarrow (Decide) \\
a^\Delta\, \neg c^\Delta & \Longrightarrow (Unit\ Propagate) \\
a^\Delta\, \neg c^\Delta\, c & \Longrightarrow (Backtrack) \\
a^\Delta\, c & \Longrightarrow (Decide) \\
a^\Delta\, c\, b^\Delta &
\end{array} \quad (2)$$

Path (1) corresponds to an execution of DPLL in the sense of (Davis et al. 1962); path (2) does not, because it applies *Decide* to $a^\Delta$ even though *Unit Propagate* could be applied in this state.

Note that the graph $\text{DP}_F$ is a modification of the *classical DPLL* graph defined in (Nieuwenhuis et al. 2006, Section 2.3). It is different in three ways. First, its states are pairs $M||F$ for all CNF formulas $F$. For the purposes of this section, it is not necessary to include $F$. Second, the description of the classical DPLL graph involves a "PureLiteral" transition rule. We dropped this rule because it does not correspond to any of the propagation rules used in answer set solvers whose algorithms we will model in this paper. Third, in the definition of that graph, each $M$ is required to be consistent. In case of DPLL, due to the simple structure of a clause, it is possible to characterize the applicability of *Backtrack* in a simple manner: when some of the clauses become inconsistent with the current partial assignment, *Backtrack* is applicable. In ASP, it is not easy to describe the applicability of *Backtrack* if only consistent states are taken into account. We introduced inconsistent states in the graph $\text{DP}_F$ to facilitate our work on extending this graph to model algorithms of answer set solvers.

In the rest of this section we give a proof of Proposition 1.

*Lemma 1*
For any CNF formula $F$ and any state $l_1 \ldots l_n$ reachable from $\emptyset$ in $\text{DP}_F$, every model $X$ of $F$ satisfies $l_i$ if it satisfies all decision literals $l_j^\Delta$ with $j \leq i$.

*Proof*

By induction on the path from $\emptyset$ to $l_1 \ldots l_n$. The property of $X$ that we need to prove trivially holds in the initial state $\emptyset$, and we will prove that all transition rules of $\text{DP}_F$ preserve it.

Take a model $X$ of $F$, and consider an edge $M \implies M'$ where $M$ is a list $l_1 \ldots l_k$ such that $X$ satisfies $l_i$ if it satisfies all decision literals $l_j^\Delta$ with $j \leq i$.

It is clear that the rule justifying the transition from $M$ to $M'$ is different from *Fail*. For each of the other three rules, $M'$ is obtained from a prefix of $M$ by appending a list of literals containing at most one decision literal. Due to the inductive hypothesis, it is sufficient to show that if $X$ satisfies all decision literals in $M'$ then $X$ satisfies all $M'$.

*Unit Propagate*: $M'$ is $M\,l$. By the inductive hypothesis, for every literal in $M$ the property in question holds. We need to show that $X \models l$. From the definition of *Unit Propagate*, for some clause $C \vee l \in F$, $\overline{C} \subseteq M$. Consequently, $M \models \neg C$. From the inductive hypothesis and the assumption that $X$ satisfies all decision literals in $M'$ and hence in $M$, it follows that $X \models M$. Since $X$ is a model of $F$, we conclude that $X \models l$.

*Decide*: $M'$ is $M\,l^\Delta$. Obvious.

*Backtrack*: $M$ has the form $P\,l^\Delta\,Q$ where $Q$ contains no decision literals. $M'$ is $P\,\overline{l}$. By the inductive hypothesis, it trivially follows that for every literal in $P$ the property in question holds. We need to show that $X \models \overline{l}$. Assume that $X \models l$. Since $Q$ does not contain decision literals, and the assumption that $X$ satisfies all decision literals in $M'$ and hence in $P$, $X$ satisfies all decision literals in $P\,l^\Delta\,Q$, that is $M$. By the inductive hypothesis, it follows that $X$ satisfies $M$. This is impossible because $M$ is inconsistent. $\square$

1Proof of Proposition 1

(a) The finiteness of $\text{DP}_F$ is obvious. For any list $N$ of literals by $|N|$ we denote the length of $N$. Any state $M$ other than *FailState* has the form $M_0\,l_1^\Delta\,M_1 \ldots l_p^\Delta\,M_p$, where $l_1^\Delta \ldots l_p^\Delta$ are all decision literals of $M$; we define $\alpha(M)$ as the sequence of nonnegative integers $|M_0|, |M_1|, \ldots, |M_p|$, and $\alpha(\textit{FailState}) = \infty$. By the definition of the transition rules defining the edges of $\text{DP}_F$, if there is an edge from a state $M$ to $M'$ in $\text{DP}_F$ then $\alpha(M) < \alpha(M')$, where $<$ is understood as the lexicographical order. It follows that if a state $M'$ is reachable from $M$ then $\alpha(M) < \alpha(M')$. Consequently the graph is acyclic.

(b) Consider any terminal state $M$ other than *FailState*. From the fact that *Decide* is not applicable, we conclude that $M$ has no unassigned literals. Since neither *Backtrack* nor *Fail* is applicable, $M$ is consistent. Consequently $M$ is an assignment. It follows that for any clause $C \vee l \in F$ if $\overline{C} \not\subseteq M$ then $C \cap M \neq \emptyset$. Furthermore, since *Unit Propagate* is not applicable, we conclude that if $\overline{C} \subseteq M$ then $l \in M$. Consequently, $M \models C \vee l$. Hence $M$ is a model of $F$.

(c) Left-to-right: Since *FailState* is reachable from $\emptyset$, there is an inconsistent state $M$ without decision literals that is reachable from $\emptyset$. By Lemma 1, any model of $F$ satisfies $M$. Since $M$ is inconsistent we conclude that $F$ has no models.

Right-to-left: From (a) it follows that there is a path from $\emptyset$ to some terminal state. By (b), this state cannot be different from *FailState*, because $F$ is unsatisfiable. $\square$

### 2.2 Logic Programs

We consider programs consisting of finitely many rules of the form

$$a \leftarrow b_1, \ldots, b_l, \textit{not } b_{l+1}, \ldots, \textit{not } b_m \qquad (3)$$

where $a$ is an atom or symbol $\bot$, and each $b_i$ ($1 \leq i \leq m$) is an atom. We will identify the body of (3) with the conjunction

$$b_1 \wedge \ldots \wedge b_l \wedge \neg b_{l+1} \wedge \ldots \neg \wedge b_m \qquad (4)$$

and also with the set of its conjunctive terms. If the head $a$ of a rule (3) is an atom then we will identify (3) with the clause

$$a \vee \neg b_1 \vee \ldots \vee \neg b_l \vee b_{l+1} \vee \ldots \vee b_m. \qquad (5)$$

If $a$ is $\bot$ then we call rule (3) a *constraint* and identify (3) with the clause

$$\neg b_1 \vee \ldots \vee \neg b_l \vee b_{l+1} \vee \ldots \vee b_m. \qquad (6)$$

We will often omit the symbol $\bot$ when referring to a constraint.

We will use two abbreviated forms for a rule (3): The first is

$$a \leftarrow B$$

where $B$ stands for $b_1, \ldots, b_l, \textit{not } b_{l+1}, \ldots, \textit{not } b_m$. The second abbreviation is

$$a \leftarrow D, F \qquad (7)$$

where $D$ stands for the *positive part of the body* $b_1, \ldots, b_l$, and $F$ stands for the *negative part of the body not* $b_{l+1}, \ldots, \textit{not } b_m$.

The *reduct* $\Pi^X$ of a program $\Pi$ with respect to a set $X$ of atoms is obtained from $\Pi$ by

- removing each rule (7) such that $\overline{F} \cap X \neq \emptyset$, and
- replacing each remaining rule (7) by $a \leftarrow D$.

A set $X$ of atoms is an *answer set* for a program $\Pi$ if $X$ is minimal (with respect to set inclusion) among the sets of atoms that satisfy the reduct $\Pi^X$ (Gelfond and Lifschitz 1988).

For example, let $\Pi$ be the program

$$\begin{aligned} a &\leftarrow \textit{not } b \\ b &\leftarrow \textit{not } a \\ c &\leftarrow a \\ d &\leftarrow d. \end{aligned} \qquad (8)$$

Consider set $\{a, c\}$. Reduct $\Pi^{\{a,c\}}$ is

$$\begin{aligned} a &\leftarrow \\ c &\leftarrow a \\ d &\leftarrow d. \end{aligned} \qquad (9)$$

Set $\{a,c\}$ satisfies the reduct and is minimal, hence $\{a,c\}$ is an answer set of $\Pi$. Consider set $\{a,c,d\}$. The reduct $\Pi^{\{a,c,d\}}$ is (9). Set $\{a,c,d\}$ satisfies the reduct but is not minimal and hence it is not an answer set of $\Pi$.

By $Bodies(\Pi, a)$ we denote the set of the bodies of all rules of $\Pi$ with head $a$. For any set $M$ of literals, by $M^+$ we denote the set of positive literals from $M$. For any consistent and complete set $M$ of literals (that is, an assignment), if $M^+$ is an answer set for a program $\Pi$, then $M$ is a model of $\Pi$. Moreover, in this case $M$ is a *supported* model of $\Pi$, in the sense that for every atom $a \in M$, $M \models B$ for some $B \in Bodies(\Pi, a)$.

A set $U$ of atoms occurring in a program $\Pi$ is said to be *unfounded* (Van Gelder et al. 1991) on a consistent set $M$ of literals w.r.t. $\Pi$ if for every $a \in U$ and every $B \in Bodies(\Pi, a)$, $\overline{B} \cap M \neq \emptyset$ or $U \cap B^+ \neq \emptyset$. There is a tight relation between unfounded sets and answer sets: For any model $M$ of a program $\Pi$, $M^+$ is an answer set for $\Pi$ if and only if $M$ contains no non-empty subsets unfounded on $M$ w.r.t. $\Pi$ (Corollary 2 from (Saccá and Zaniolo 1990)[9]).

For instance, let $\Pi$ be program (8) and let $M$ be a consistent set $\{a, \neg b, c, d\}$ of literals. We already demonstrated that $M^+ = \{a,c,d\}$ is not an answer set of $\Pi$. Accordingly, its subset $\{d\}$ is unfounded on $\{a, \neg b, c, d\}$ w.r.t. $\Pi$, because the only rule in $\Pi$ with $d$ in the head

$$d \leftarrow d$$

is such that $U \cap B^+ = \{d\} \cap \{d\} \neq \emptyset$.

We say that a program $\Pi$ *entails* a formula $F$ when for any consistent and complete set $M$ of literals, if $M^+$ is an answer set for $\Pi$, then $M \models F$. For instance, any program $\Pi$ entails each rule occurring in $\Pi$.

## 3 Generating Supported Models

In Section 4 we will define, for an arbitrary program $\Pi$, a graph $\text{SM}_\Pi$ representing the application of the SMODELS algorithm to $\Pi$; the terminal nodes of $\text{SM}_\Pi$ are answer sets of $\Pi$. As a step in this direction, we describe here a simpler graph ATLEAST$_\Pi$.

### 3.1 Graph ATLEAST$_\Pi$

The terminal nodes of ATLEAST$_\Pi$ are supported models of $\Pi$. The transition rules defining ATLEAST$_\Pi$ are closely related to procedure *Atleast* (Simons 2000, Sections 4.1), which is one of the core procedures of the SMODELS algorithm.

The nodes of ATLEAST$_\Pi$ are the states relative to the set of atoms occurring in $\Pi$. The edges of the graph ATLEAST$_\Pi$ are described by the transition rules *Decide*, *Fail*, *Backtrack* introduced in Section 2.1 and the additional transition rules[10] presented

---

[9] The Corollary 2 from (Saccá and Zaniolo 1990) refers to "assumption sets" rather than unfounded sets. But as the authors noted, in the context of this corollary the two concepts are equivalent.

[10] The names of some of these rules follow (Ward 2004).

*Unit Propagate LP*:
$$M \Longrightarrow M\,a \quad \text{if} \quad \begin{cases} a \leftarrow B \in \Pi \text{ and} \\ B \subseteq M \end{cases}$$

*All Rules Cancelled*:
$$M \Longrightarrow M\,\neg a \quad \text{if } \overline{B} \cap M \neq \emptyset \text{ for all } B \in \textit{Bodies}(\Pi, a)$$

*Backchain True*:
$$M \Longrightarrow M\,l \quad \text{if} \quad \begin{cases} a \leftarrow B \in \Pi, \\ a \in M, \\ \overline{B'} \cap M \neq \emptyset \text{ for all } B' \in \textit{Bodies}(\Pi, a) \setminus \{B\}, \\ l \in B \end{cases}$$

*Backchain False*:
$$M \Longrightarrow M\,\overline{l} \quad \text{if} \quad \begin{cases} a \leftarrow l, B \in \Pi, \\ \neg a \in M \text{ or } a = \bot, \\ B \subseteq M \end{cases}$$

Fig. 2. The additional transition rules of the graph ATLEAST$_\Pi$.

in Figure 2. Note that each of the rules *Unit Propagate LP* and *Backchain False* is similar to *Unit Propagate*: the former corresponds to *Unit Propagate* on $C \vee l$ where $l$ is the head of the rule, and the latter corresponds to *Unit Propagate* on $C \vee l$ where $\overline{l}$ is an element of the body of the rule.

This graph can be used for deciding whether program $\Pi$ has a supported model by constructing a path from $\emptyset$ to a terminal node:

*Proposition 2*
For any program $\Pi$,

(a) graph ATLEAST$_\Pi$ is finite and acyclic,
(b) any terminal state of ATLEAST$_\Pi$ other than *FailState* is a supported model of $\Pi$,
(c) *FailState* is reachable from $\emptyset$ in ATLEAST$_\Pi$ if and only if $\Pi$ has no supported models.

For instance, let $\Pi$ be program (8). Here is a path in ATLEAST$_\Pi$:

$$\begin{array}{lll} \emptyset & \Longrightarrow & (\textit{Decide}) \\ a^\Delta & \Longrightarrow & (\textit{Unit Propagate LP}) \\ a^\Delta\,c & \Longrightarrow & (\textit{All Rules Cancelled}) \\ a^\Delta\,c\,\neg b & \Longrightarrow & (\textit{Decide}) \\ a^\Delta\,c\,\neg b\,d^\Delta & & \end{array} \qquad (10)$$

Since the state $a^\Delta\,c\,\neg b\,d^\Delta$ is terminal, Proposition 2(b) asserts that $\{a, c, \neg b, d\}$ is a supported model of $\Pi$.

The assertion of Proposition 2 will remain true if we drop the transition rules *Backchain True* and *Backchain False* from the definition of ATLEAST$_\Pi$.

In the rest of this section we give a proof of Proposition 2.

*Lemma 2*

For any program $\Pi$ and any state $l_1 \ldots l_n$ reachable from $\emptyset$ in ATLEAST$_\Pi$, every supported model $X$ for $\Pi$ satisfies $l_i$ if it satisfies all decision literals $l_j^\Delta$ with $j \leq i$.

*Proof*

By induction on the path from $\emptyset$ to $l_1 \ldots l_n$. Similar to the proof of Lemma 1. We will show that the property in question is preserved when the transition from $M$ to $M'$ is justified by any of the four new rules.

Take a supported model $X$ for $\Pi$, and consider an edge $M \Longrightarrow M'$ where $M$ is a list $l_1 \ldots l_k$ such that $X$ satisfies $l_i$ if it satisfies all decision literals $l_j^\Delta$ with $j \leq i$.

Assume that $X$ satisfies all decision literals in $M'$.

*Unit Propagate LP*: $M'$ is $M\,a$. By the inductive hypothesis, for every literal in $M$ the property in question holds. We need to show that $X \models a$. By the definition of *Unit Propagate LP*, $B \subseteq M$ for some rule $a \leftarrow B$. Consequently, $M \models B$. From the inductive hypothesis and the assumption that $X$ satisfies all decision literals in $M'$ and hence in $M$, it follows that $X \models M$. Since $X$ is a model of $\Pi$ we conclude that $X \models a$.

*All Rules Cancelled*: $M'$ is $M\,\neg a$ and $\overline{B} \cap M \neq \emptyset$ for every $B \in \mathit{Bodies}(\Pi, a)$. Consequently, $M \models \neg B$ for every $B \in \mathit{Bodies}(\Pi, a)$. By the inductive hypothesis, for every literal in $M$ the property in question holds. We need to show that $X \models \neg a$. By contradiction. Assume that $X \models a$. From the inductive hypothesis and the assumption that $X$ satisfies all decision literals in $M'$ and hence in $M$, it follows that $X \models M$. Since $M \models \neg B$ for every $B \in \mathit{Bodies}(\Pi, a)$, it follows that $X \models \neg B$. We conclude that $X$ is not a supported model of $\Pi$.

*Backchain True*: $M'$ is $M\,l$. By the inductive hypothesis, for every literal in $M$ the property in question holds. We need to show that $X \models l$. By contradiction. Assume $X \models \overline{l}$. Consider the rule $a \leftarrow B$ corresponding to this application of *Backchain True*. Since $l \in B$, $X \models \neg B$. By the definition of *Backchain True*, $\overline{B'} \cap M \neq \emptyset$ for every $B'$ in $\mathit{Bodies}(\Pi, a) \setminus B$. Consequently, $M \models \neg B'$ for every $B'$ in $\mathit{Bodies}(\Pi, a) \setminus B$. From the inductive hypothesis and the assumption that $X$ satisfies all decision literals in $M'$ and hence in $M$, it follows that $X \models M$. We conclude that $X \models \neg B'$ for every $B'$ in $\mathit{Bodies}(\Pi, a) \setminus B$. Hence $X$ is not supported by $\Pi$.

*Backchain False*: $M'$ is $M\,\overline{l}$. By the inductive hypothesis, for every literal in $M$ the property in question holds. We need to show that $X \models \overline{l}$. By contradiction. Assume that $X \models l$. By the definition of *Backchain False* there exists a rule $a \leftarrow l, B$ in $\Pi$ such that $\neg a \in M$ and $B \subseteq M$. Consequently, $M \models \neg a$ and $M \models B$. From the inductive hypothesis and the assumption that $X$ satisfies all decision literals in $M'$ and hence in $M$, it follows that $X \models M$. We conclude that $X \models \neg a$ and $X \models B$. From the fact that $X \models l$, it follows that $X$ does not satisfy the rule $a \leftarrow l, B$, so that it is not a model of $\Pi$. □

1Proof of Proposition 2

Parts (a) and (c) are proved as in the proof of Proposition 1, using Lemma 2.

(b) Let $M$ be a terminal state so that none of the rules are applicable. From the

fact that *Decide* is not applicable, we conclude that $M$ assigns all literals. Since neither *Backtrack* nor *Fail* is applicable, $M$ is consistent. Consequently, $M$ is an assignment. Since *Unit Propagate LP* is not applicable, it follows that for every rule $a \leftarrow B \in \Pi$, if $B \subseteq M$ then $a \in M$. Consequently, if $M \models B$ then $M \models a$. We conclude that $M$ is a model of $\Pi$. We will now show that $M$ is a supported model of $\Pi$. By contradiction. Suppose that $M$ is not a supported model. Then, there is an atom $a \in M$ such that $M \not\models B$ for every $B \in Bodies(\Pi, a)$. Since $M$ is consistent, $\overline{B} \cap M \neq \emptyset$ for every $B \in Bodies(\Pi, a)$. Consequently, *All Rules Cancelled* is applicable. This contradicts the assumption that $M$ is terminal. □

The fact that the assertion of Proposition 2 remains true if we drop the transition rules *Backchain True* and *Backchain False* from the definition of ATLEAST$_\Pi$ follows from the proof of Proposition 2 (b) that does not refer to those rules.

### 3.2 Relation between DP$_F$ and ATLEAST$_\Pi$

It is well known that the supported models of a program can be characterized as models of program's completion in the sense of (Clark 1978). It turns out that the graph ATLEAST$_\Pi$ is identical to the graph DP$_F$, where $F$ is the (clausified) completion of $\Pi$. To make this claim precise, we first review the notion of completion.

For any program $\Pi$, its completion consists of $\Pi$ and the formulas that can be written as

$$\neg a \vee \bigvee_{B \in Bodies(\Pi, a)} B \tag{11}$$

for every atom $a$ in $\Pi$. $\iota CNF - Comp(\Pi)$ is the completion converted to CNF using straightforward equivalent transformations. In other words, $\iota CNF - Comp(\Pi)$ consists of clauses of two kinds:

1. the rules $a \leftarrow B$ of the program written as clauses

$$a \vee \overline{B}, \tag{12}$$

2. formulas (11) converted to CNF using the distributivity of disjunction over conjunction[11].

*Proposition 3*
For any program $\Pi$, the graphs ATLEAST$_\Pi$ and DP$_{CNF\text{-}Comp(\Pi)}$ are equal.

For instance, let $\Pi$ be the program

$$\begin{aligned} &a \leftarrow b, \ not \ c \\ &b. \end{aligned} \tag{13}$$

Its completion is

$$(a \leftrightarrow b \wedge \neg c) \wedge b \wedge \neg c, \tag{14}$$

---

[11] It is essential that repetitions are not removed in the process of clausification. For instance, $\iota CNF - Comp(a \leftarrow not \ a)$ is the formula $(a \vee a) \wedge (\neg a \vee \neg a)$.

and $_1CNF - Comp(\Pi)$ is

$$(a \vee \neg b \vee c) \wedge (\neg a \vee b) \wedge (\neg a \vee \neg c) \wedge b \wedge \neg c. \tag{15}$$

Proposition 3 asserts that $\text{ATLEAST}_\Pi$ coincides with $\text{DP}_{CNF\text{-}Comp(\Pi)}$.

From Proposition 3, it follows that applying the *Atleast* algorithm to a program essentially amounts to applying DPLL to its completion.

In the rest of this section we give a proof of Proposition 3.

It is easy to see that the states of the graphs $\text{ATLEAST}_\Pi$ and $\text{DP}_{CNF\text{-}Comp(\Pi)}$ coincide. We will now show that the edges of $\text{ATLEAST}_\Pi$ and $\text{DP}_{CNF\text{-}Comp(\Pi)}$ coincide also.

It is clear that there is an edge $M \Longrightarrow M'$ in $\text{ATLEAST}_\Pi$ justified by the rule *Decide* if and only if there is an edge $M \Longrightarrow M'$ in $\text{DP}_{CNF\text{-}Comp(\Pi)}$ justified by *Decide*. The same holds for the transition rules *Fail* and *Backtrack*.

We will now show that if there is an edge from a state $M$ to a state $M'$ in the graph $\text{DP}_{CNF\text{-}Comp(\Pi)}$ justified by the transition rule *Unit Propagate* then there is an edge from $M$ to $M'$ in $\text{ATLEAST}_\Pi$. Consider a clause $C \vee l \in {_1CNF} - Comp(\Pi)$ such that $\overline{C} \subseteq M$. We will consider two cases, depending on whether $C \vee l$ comes from (12) or from the CNF of (11).

Case 1. $C \vee l$ is $a \vee \overline{B}$ corresponding to a rule $a \leftarrow B$.

Case 1.1. $l$ is $a$. Then there is an edge from $M$ to $M'$ in $\text{ATLEAST}_\Pi$ justified by the transition rule *Unit Propagate LP*.

Case 1.2. $l$ is an element of $\overline{B}$. Then $B$ has the form $\overline{l}, D$ and $C$ is $a \vee \overline{D}$. From $\overline{C} \subseteq M$ we conclude that $D \subseteq M$ and $\neg a \in M$. There is an edge from $M$ to $M'$ in the graph $\text{ATLEAST}_\Pi$ justified by the following instance of *Backchain False*:

$$M \Longrightarrow M\, l \quad \text{if} \quad \begin{cases} a \leftarrow \overline{l}, D \ \in\ \Pi, \\ \neg a \in M, \\ D \subseteq M. \end{cases}$$

Case 2. $C \vee l$ has the form $\neg a \vee D$, where $D$ is one of the clauses of the CNF of

$$\bigvee_{B \in Bodies(\Pi, a)} B.$$

Then $D$ has the form

$$\bigvee_{B \in Bodies(\Pi, a)} f(B)$$

where $f$ is a function that maps every $B \in Bodies(\Pi, a)$ to an element of $B$.

Case 2.1. $l$ is $\neg a$. Then $C$ is $D$, so that $\overline{D} \subseteq M$. Consequently $\overline{f(B)} \in \overline{B} \cap \overline{D} \subseteq \overline{B} \cap M$, so that $\overline{B} \cap M \neq \emptyset$ for every $B \in Bodies(\Pi, a)$. There is an edge from $M$ to $M'$ in $\text{ATLEAST}_\Pi$ justified by *All Rules Cancelled*.

Case 2.2. $l$ is an element of $D$. From the construction of $D$, it follows that $l = f(B) \in B$ for some rule $a \leftarrow B$. Then $C$ is

$$\neg a \vee \bigvee_{B' \in Bodies(\Pi, a) \setminus B} f(B').$$

From $\overline{C} \subseteq M$ we conclude that $a \in M$ and that $\overline{f(B')} \in M$ for every $B' \in$

$Bodies(\Pi, a) \setminus B$. Since $f(B')$ is a conjunctive term of $B'$, it follows that $\overline{B'} \cap M \neq \emptyset$. Then there is an edge from $M$ to $M'$ in ATLEAST$_\Pi$ justified by *Backchain True*.

We will now show that if there is an edge from a state $M$ to a state $M'$ in the graph ATLEAST$_\Pi$ justified by one of the transition rules *Unit Propagate LP*, *All Rules Cancelled*, *Backchain True*, and *Backchain False* then there is an edge from $M$ to $M'$ in DP$_{CNF\text{-}Comp(\Pi)}$.

Case 1. The edge is justified by *Unit Propagate LP*. Then there is a rule $a \leftarrow B \in \Pi$ where $B \subseteq M$, and $M'$ is $M\,a$. By the construction of $1CNF - Comp(\Pi)$, $a \vee \overline{B} \in 1CNF - Comp(\Pi)$. There is an edge from $M$ to $M'$ in DP$_{CNF\text{-}Comp(\Pi)}$ justified by the following instance of *Unit Propagate*:

$$M \Longrightarrow M\,a \text{ if } \begin{cases} \overline{B} \vee a \in 1CNF - Comp(\Pi) \text{ and} \\ B \subseteq M. \end{cases}$$

Case 2. The edge is justified by *All Rules Cancelled*. By the definition of *All Rules Cancelled*, there is an atom $a$ such that for all $B \in Bodies(\Pi, a)$, $\overline{B} \cap M \neq \emptyset$; and $M'$ is $M\,\neg a$. Consequently, $M$ contains the complement of some literal in $B$. Denote one of such literals by $f(B)$, so that $\overline{f(B)} \in M$. From the construction of $1CNF - Comp(\Pi)$,

$$\neg a \vee \bigvee_{B \in Bodies(\Pi, a)} f(B)$$

belongs to $1CNF - Comp(\Pi)$. By the choice of $f$,

$$\overline{\bigvee_{B \in Bodies(\Pi, a)} f(B)} \subseteq M.$$

There is an edge from $M$ to $M'$ in DP$_{CNF\text{-}Comp(\Pi)}$ justified by the following instance of *Unit Propagate*:

$$M \Longrightarrow M\,\neg a \text{ if } \begin{cases} \bigvee_{B \in Bodies(\Pi, a)} f(B) \vee \neg a \in 1CNF - Comp(\Pi), \\ \\ \overline{\bigvee_{B \in Bodies(\Pi, a)} f(B)} \subseteq M. \end{cases}$$

Case 3. The edge is justified by *Backchain True*. By the definition of *Backchain True*, there is a rule $a \leftarrow B \in \Pi$ and a literal $l \in B$ such that $a \in M$; for all $B' \in Bodies(\Pi, a) \setminus B$, $\overline{B'} \cap M \neq \emptyset$; and $M'$ is $M\,l$. Let $f(B')$ be an element of $B'$ such that $\overline{f(B')} \in M$. From the construction of $1CNF - Comp(\Pi)$,

$$\neg a \vee l \vee \bigvee_{B' \in Bodies(\Pi, a) \setminus B} f(B')$$

belongs to $1CNF - Comp(\Pi)$. By the choice of $f$,

$$\overline{\bigvee_{B' \in Bodies(\Pi, a) \setminus B} f(B')} \subseteq M.$$

There is an edge from $M$ to $M'$ in DP$_{CNF\text{-}Comp(\Pi)}$ justified by the following instance

of *Unit Propagate*:

$$M \Longrightarrow M\,l \quad \text{if} \quad \begin{cases} \neg a \vee l \vee \bigvee_{B' \in Bodies(\Pi,a)\setminus B} f(B') \in {}_1CNF - Comp(\Pi), \\ \overline{(\neg a \vee \bigvee_{B' \in Bodies(\Pi,a)\setminus B} f(B'))} \subseteq M. \end{cases}$$

Case 4. The edge is justified by *Backchain False*. By the definition of *Backchain False*, there is a rule $a \leftarrow l, B \in \Pi$ such that $\neg a \in M$, $B \subseteq M$, and $M'$ is $M\,\overline{l}$. By the construction of $_1CNF - Comp(\Pi)$, $a \vee \overline{B} \vee \overline{l} \in {}_1CNF - Comp(\Pi)$. There is an edge from $M$ to $M'$ in DP$_{CNF\text{-}Comp(\Pi)}$ justified by the following instance of *Unit Propagate*:

$$M \Longrightarrow M\,\overline{l} \quad \text{if} \quad \begin{cases} a \vee \overline{B} \vee \overline{l} \in {}_1CNF - Comp(\Pi) \text{ and} \\ \overline{a \vee \overline{B}} \subseteq M. \end{cases}$$

□

## 4 Answer Set Solver Smodels

### 4.1 Abstract Smodels

We now describe the graph SM$_\Pi$ that represents the application of the SMODELS algorithm to program $\Pi$. SM$_\Pi$ is a graph whose nodes are the same as the nodes of the graph ATLEAST$_\Pi$. The edges of SM$_\Pi$ are described by the transition rules of ATLEAST$_\Pi$ and the additional transition rule:

*Unfounded*:
$$M \Longrightarrow M\neg a \quad \text{if} \quad \begin{cases} M \text{ is consistent, and} \\ a \in U \text{ for a set } U \text{ unfounded on } M \text{ w.r.t. } \Pi. \end{cases}$$

This transition rule of SM$_\Pi$ is closely related to procedure *Atmost* (Simons 2000, Sections 4.2), which together with the procedure *Atleast* forms the core of the SMODELS algorithm.

The graph SM$_\Pi$ can be used for deciding whether program $\Pi$ has an answer set by constructing a path from $\emptyset$ to a terminal node:

*Proposition 4*
For any program $\Pi$,

(a) graph SM$_\Pi$ is finite and acyclic,
(b) for any terminal state $M$ of SM$_\Pi$ other than *FailState*, $M^+$ is an answer set of $\Pi$,
(c) *FailState* is reachable from $\emptyset$ in SM$_\Pi$ if and only if $\Pi$ has no answer sets.

To illustrate the difference between SM$_\Pi$ and ATLEAST$_\Pi$, assume again that $\Pi$ is program (8). Path (10) in the graph ATLEAST$_\Pi$ is also a path in SM$_\Pi$. But state $a^\Delta c\neg b\,d^\Delta$, which is terminal in ATLEAST$_\Pi$, is not terminal in SM$_\Pi$. This is not surprising, since $\{a, c, \neg b, d\}^+ = \{a, c, d\}$ is not an answer set of $\Pi$. To get to a state that is terminal in $\text{SM}_\Pi$, we need two more steps:

$$
\begin{aligned}
&\vdots \\
&a^\Delta\, c\, \neg b\, d^\Delta &&\Longrightarrow (\textit{Unfounded},\ U = \{d\}) \\
&a^\Delta\, c\, \neg b\, d^\Delta\, \neg d &&\Longrightarrow (\textit{Backtrack}) \\
&a^\Delta\, c\, \neg b\, \neg d
\end{aligned}
\qquad (16)
$$

Proposition 4(b) asserts that $\{a, c\}$ is an answer set of $\Pi$.

The assertion of Proposition 4 will remain true if we drop the transition rules *All Rules Cancelled*, *Backchain True*, and *Backchain False* from the definition of $\text{SM}_\Pi$.

In the rest of this section we give a proof of Proposition 4.

We say that a model $M$ of a program $\Pi$ is *unfounded-free* if no non-empty subset of $M$ is an unfounded set on $M$ w.r.t. $\Pi$.

*Lemma 3 (Corollary 2 from (Saccá and Zaniolo 1990))*
For any model $M$ of a program $\Pi$, $M^+$ is an answer set for $\Pi$ if and only if $M$ is unfounded-free.

*Lemma 4*
For any unfounded set $U$ on a consistent set $M$ of literals w.r.t. a program $\Pi$, and any assignment $X$, if $X \models M$ and $X \cap U \neq \emptyset$, then $X^+$ is not an answer set for $\Pi$.

*Proof*
Assume that $X^+$ is an answer set for $\Pi$. Then $X$ is a model of $\Pi$. By Lemma 3, it follows that $X$ is unfounded-free. Hence any non-empty subset of $X$ including $X \cap U$ is not unfounded on $X$. This means that for some rule $a \leftarrow B$ in $\Pi$ such that $a \in X \cap U$, $\overline{B} \cap X = \emptyset$ and $X \cap U \cap B^+ = \emptyset$. From $X \models M$ ($M \subseteq X$) and $\overline{B} \cap X = \emptyset$ we conclude that $\overline{B} \cap M = \emptyset$. Since $\overline{B} \cap X = \emptyset$ and $X$ is an assignment, $B \subseteq X$. It follows that $B^+ \subseteq X$. Consequently $U \cap B^+ = X \cap U \cap B^+ = \emptyset$. This contradicts the assumption that $U$ is an unfounded set on $M$. □

*Lemma 5*
For any program $\Pi$, any state $l_1 \ldots l_n$ reachable from $\emptyset$ in $\text{SM}_\Pi$, and any assignment $X$, if $X^+$ is an answer set for $\Pi$ then $X$ satisfies $l_i$ if it satisfies all decision literals $l_j^\Delta$ with $j \leq i$.

*Proof*
By induction on the path from $\emptyset$ to $l_1 \ldots l_n$. Recall that for any assignment $X$, if $X^+$ is an answer set for $\Pi$, then $X$ is a supported model of $\Pi$, and that the transition system $\text{SM}_\Pi$ extends $\text{ATLEAST}_\Pi$ only by the transition rule *Unfounded*. Given our proof of Lemma 2, we only need to demonstrate that application of *Unfounded* preserves the property.

Consider a transition $M \Longrightarrow M'$ justified by *Unfounded*, where $M$ is a sequence $l_1 \ldots l_k$. $M'$ is $M \neg a$, such that $a \in U$, where $U$ is an unfounded set on $M$ w.r.t $\Pi$. Take any assignment $X$ such that $X^+$ is an answer set for $\Pi$ and $X$ satisfies all decision literals $l_j^\Delta$ with $j \leq k$. By the inductive hypothesis, $X \models M$. Then $X \models \neg a$. Indeed, otherwise $a$ would be a common element of $X$ and $U$, and $X \cap U$ would be non-empty, which contradicts Lemma 4.  □

₁Proof of Proposition 4
Parts (a) and (c) are proved as in the proof of Proposition 1, using Lemma 5.
(b) As in the proof of Proposition 2(b) we conclude that $M$ is a model of $\Pi$. Assume that $M^+$ is not an answer set. Then, by Lemma 3, there is a non-empty unfounded set $U$ on $M$ w.r.t. $\Pi$ such that $U \subseteq M$. It follows that *Unfounded* is applicable (with an arbitrary $a \in U$). This contradicts the assumption that $M$ is terminal. □

The fact that the assertion of Proposition 4 remains true if we drop the transition rules *All Rules Cancelled*, *Backchain True*, and *Backchain False* from the definition of SM$_\Pi$ follows from the proof of Proposition 4 (b) that does not refer to those rules.

### 4.2 Smodels Algorithm

We can view a path in the graph SM$_\Pi$ as a description of a process of search for an answer set for a program $\Pi$ by applying inference rules. Therefore, we can characterize the algorithm of an answer set solver that utilizes the inference rules of SM$_\Pi$ by describing a strategy for choosing a path in SM$_\Pi$. A strategy can be based, in particular, on assigning priorities to some or all inference rules of SM$_\Pi$, so that a solver will never apply a transition rule in a state if a rule with higher priority is applicable to the same state.

We use this method to describe the SMODELS algorithm. System SMODELS assigns priorities to the inference rules of SM$_\Pi$ as follows:

*Backtrack, Fail* ≫
*Unit Propagate LP, All Rules Cancelled, Backchain True, Backchain False* ≫
*Unfounded* ≫
*Decide*.

For example, let $\Pi$ be program (8). The SMODELS algorithm may follow a path

$$\begin{array}{ll}
\emptyset & \Longrightarrow (\textit{Decide}) \\
a^\Delta & \Longrightarrow (\textit{Unit Propagate LP}) \\
a^\Delta c & \Longrightarrow (\textit{All Rules Cancelled}) \\
a^\Delta c \neg b & \Longrightarrow (\textit{Unfounded}) \\
a^\Delta c \neg b \neg d &
\end{array}$$

in the graph SM$_\Pi$, whereas it may never follow path (10), because *Unfounded* has a higher priority than *Decide*.

### 4.3 Tight Programs

We will now review the definitions of a positive dependency graph and a tight program. The *positive dependency graph* of a program $\Pi$ is the directed graph $G$ such that

- the nodes of $G$ are the atoms occurring in $\Pi$, and
- $G$ contains the edges from $a$ to $b_i$ ($1 \leq i \leq l$) for each rule

$$a \leftarrow b_1, \ldots, b_l, \mathit{not}\ b_{l+1}, \ldots, \mathit{not}\ b_m$$

in $\Pi$ where $a$ is an atom.

A program is *tight* if its positive dependency graph is acyclic. For instance, program (8) is not tight since its positive dependency graph has a cycle due to the rule $d \leftarrow d$. On the other hand, the program constructed from (8) by removing this rule is tight.

Recall that for any program $\Pi$ and any assignment $M$, if $M^+$ is an answer set of $\Pi$ then $M$ is a supported model of $\Pi$. For the case of tight programs, the converse holds also: $M^+$ is an answer set for $\Pi$ if and only if $M$ is a supported model of $\Pi$ (Fages 1994) or, in other words, is a model of the completion of $\Pi$.

It turns out that for tight programs the graph $\text{SM}_\Pi$ is "almost identical" to the graph $\text{DP}_F$, where $F$ is the clausified completion of $\Pi$. To make this claim precise, we need the following terminology.

We say that an edge $M \Longrightarrow M'$ in the graph $\text{SM}_\Pi$ is *singular* if

- the only transition rule justifying this edge is *Unfounded*, and
- some edge $M \Longrightarrow M''$ can be justified by a transition rule other than *Unfounded* or *Decide*.

For instance, let $\Pi$ be the program

$$\begin{aligned} a &\leftarrow b \\ b &\leftarrow c. \end{aligned}$$

The edge

$$\begin{aligned} a^\Delta\, b^\Delta\, \neg c^\Delta \quad &\Longrightarrow (\mathit{Unfounded},\ U = \{a, b\}) \\ a^\Delta\, b^\Delta\, \neg c^\Delta\, \neg a \end{aligned}$$

in the graph $\text{SM}_\Pi$ is singular, because the edge

$$\begin{aligned} a^\Delta\, b^\Delta\, \neg c^\Delta \quad &\Longrightarrow (\mathit{All\ Rules\ Cancelled}) \\ a^\Delta\, b^\Delta\, \neg c^\Delta\, \neg b \end{aligned}$$

belongs to $\text{SM}_\Pi$ also.

With respect to the actual SMODELS algorithm (Simons 2000), singular edges of the graph $\text{SM}_\Pi$ are inessential: in view of priorities for choosing a path in $\text{SM}_\Pi$ described in Section 4.2 SMODELS never follows a singular edge. Indeed, the transition rule *Unfounded* has the lower priority than any other transition rule but *Decide*. By $\text{SM}_\Pi^-$ we denote the graph obtained from $\text{SM}_\Pi$ by removing all singular edges.

*Proposition 5*
For any tight program $\Pi$, the graph $\text{SM}_\Pi^-$ is equal to each of the graphs $\text{ATLEAST}_\Pi$ and $\text{DP}_{\mathit{CNF\text{-}Comp}(\Pi)}$.

For instance, let $\Pi$ be the program (13). This program is tight, its completion is (14), and $\mathit{1CNF} - \mathit{Comp}(\Pi)$ is formula (15). Proposition 5 asserts that, $\text{SM}_\Pi^-$ coincides with $\text{DP}_{\mathit{CNF\text{-}Comp}(\Pi)}$ and with $\text{ATLEAST}_\Pi$.

From Proposition 5, it follows that applying the SMODELS algorithm to a tight program essentially amounts to applying DPLL to its completion. A similar relationship, in terms of pseudocode representations of SMODELS and DPLL, is established in (Giunchiglia and Maratea 2005).

In the rest of this section we give a proof of Proposition 5.

*Lemma 6*

For any tight program $\Pi$ and any non-empty unfounded set $U$ on a consistent set $M$ of literals w.r.t. $\Pi$ there is an atom $a \in U$ such that for every $B \in Bodies(\Pi, a)$, $\overline{B} \cap M \neq \emptyset$.

*Proof*

By contradiction. Assume that, for every $a \in U$ there exists $B \in Bodies(\Pi, a)$ such that $\overline{B} \cap M = \emptyset$. By the definition of an unfounded set it follows that for every atom $a \in U$ there is $B \in Bodies(\Pi, a)$ such that $U \cap B^+ \neq \emptyset$. Consequently the subgraph of the positive dependency graph of $\Pi$ induced by $U$ has no terminal nodes. Then, the program $\Pi$ is not tight. □

1Proof of Proposition 5

In view of Proposition 3, it is sufficient to prove that $SM_\Pi^-$ equals $ATLEAST_\Pi$; or, in other words, that every edge of $SM_\Pi$ justified by the rule *Unfounded* only is singular. Consider such an edge $M \Longrightarrow M'$. We need to show that some transition rule other than *Unfounded* or *Decide* is applicable to $M$. By the definition of *Unfounded*, $M$ is consistent and there exists a non-empty set $U$ unfounded on $M$ w.r.t. $\Pi$. By Lemma 6, it follows that there is an atom $a \in U$ such that for every $B \in Bodies(\Pi, a)$, $\overline{B} \cap M \neq \emptyset$. Therefore, the transition rule *All Rules Cancelled* is applicable to $M$. □

## 5 Generate and Test

In this section, we present a modification of the graph $DP_F$ (Section 2.1) that includes testing "partial" assignments of $F$ found by DPLL.

Let $F$ be a CNF formula, and let $G$ be a formula formed from atoms occurring in $F$. The terminal nodes of the graph $GT_{F,G}$ defined below are models of formula $F \wedge G$.

This modification of the graph $DP_F$ is of interest, for example, in connection with the fact that answer sets of a program $\Pi$ can be characterized as models of its completion extended by so called loop formulas of $\Pi$ (Lin and Zhao 2002). If $1CNF - Comp(\Pi)$, as above, is the completion converted to CNF, and $LF(\Pi)$ is the conjunction of all loop formulas of $\Pi$, then for any assignment $M$, $M^+$ is an answer set of $\Pi$ iff $M$ is a model of $1CNF - Comp(\Pi) \wedge LF(\Pi)$. Hence, the terminal nodes of the graph $GT_{CNF\text{-}Comp(\Pi), LF(\Pi)}$ will correspond to answer sets of $\Pi$.

The nodes of the graph $GT_{F,G}$ are the same as the nodes of the graph $DP_F$. The edges of $GT_{F,G}$ are described by the transition rules of $DP_F$ and the additional

transition rule:

*Test*:
$$M \implies M\bar{l} \text{ if } \begin{cases} M \text{ is consistent,} \\ G \models \overline{M}, \\ l \in M \end{cases}$$

It is easy to see that the graph $\text{DP}_F$ is a subgraph of $\text{GT}_{F,G}$. The latter graph can be used for deciding whether a formula $F \wedge G$ has a model by constructing a path from $\emptyset$ to a terminal node:

*Proposition 6*
For any CNF formula $F$ and a formula $G$ formed from atoms occurring in $F$,

(a) graph $\text{GT}_{F,G}$ is finite and acyclic,
(b) any terminal state of $\text{GT}_{F,G}$ other than *FailState* is a model of $F \wedge G$,
(c) *FailState* is reachable from $\emptyset$ in $\text{GT}_{F,G}$ if and only if $F \wedge G$ is unsatisfiable.

Note that to verify the applicability of the new transition rule *Test* we need a procedure for testing whether $G$ entails a clause, but there is no need to explicitly write out $G$. This is important because $LF(\Pi)$ can be very long (Lin and Zhao 2002).

For instance, let $\Pi$ be the nontight program

$$d \leftarrow d.$$

Its completion is

$$d \leftrightarrow d,$$

and $1CNF - Comp(\Pi)$ is

$$(d \vee \neg d).$$

This program has one loop formula

$$d \to \bot.$$

Proposition 6 asserts that a terminal state $\neg d$ of $\text{GT}_{CNF\text{-}Comp(\Pi),\, d \to \bot}$ is a model of $1CNF - Comp(\Pi) \wedge LF(\Pi)$. It follows that $\{\neg d\}^+ = \emptyset$ is an answer set of $\Pi$. To compare with the graph $\text{DP}_{CNF\text{-}Comp(\Pi)}$: state $d$ is a terminal state in $\text{DP}_{CNF\text{-}Comp(\Pi)}$ whereas $d$ is not a terminal state in $\text{GT}_{CNF\text{-}Comp(\Pi),\, d \to \bot}$ because the transition rule *Test* is applicable to this state.

ASP-SAT with Backtracking (Giunchiglia et al. 2006) is a procedure that computes models of the completion of the given program using DPLL, and tests them until an answer set is found. The application of this procedure to a program $\Pi$ can be viewed as constructing a path from $\emptyset$ to a terminal node in the graph $\text{GT}_{CNF\text{-}Comp(\Pi),\, LF(\Pi)}$ by adopting a strategy that *Test* is applied to a state $M$ only when $M$ is an assignment.

In the rest of this section we give a proof of Proposition 6.

*Lemma 7*
For any CNF formula $F$, a formula $G$ formed from atoms occurring in $F$, and a path from $\emptyset$ to a state $l_1 \ldots l_n$ in $\text{GT}_{F,G}$, any model $X$ of $F \wedge G$ satisfies $l_i$ if it satisfies all decision literals $l_j^\Delta$ with $j \leq i$.

*Proof*
By induction on the path from $\emptyset$ to $l_1 \ldots l_n$. Similar to the proof of Lemma 1. We will show that the property in question is preserved by the transition rule *Test*.

Take a model $X$ of $F \wedge G$ and consider an edge $M \Longrightarrow M'$ where $M$ is a list $l_1 \ldots l_k$ such that $X$ satisfies $l_i$ if it satisfies all decision literals $l_j^\Delta$ with $j \leq i$.

Assume that $X$ satisfies all decision literals from $M$. By the inductive hypothesis, $X \models M$. We will show that the rule justifying the transition from $M$ to $M'$ is different from *Test*. By contradiction. $M'$ is $M\,\overline{l}$. By the definition of *Test*, $G \models \overline{M}$. Since $X$ is a model of $F \wedge G$ it follows that $X \models \overline{M}$. This contradicts the fact that $X \models M$. □

ıProof of Proposition 6
Part (a) and part (c) Right-to-left are proved as in the proof of Proposition 1.
(b) Let $M$ be any terminal state other than *FailState*. As in the proof of Proposition 1(b) it follows that $M$ is a model of $F$. The transition rule *Test* is not applicable. Hence $G \not\models \overline{M}$. In other words $M$ is a model of $G$. We conclude that $M$ is a model of $F \wedge G$
(c) Left-to-right: Since *FailState* is reachable from $\emptyset$, there is a state $M$ without decision literals such that $M$ is reachable from $\emptyset$ and the transition rule *Fail* is applicable in $M$. Then, $M$ is inconsistent. By Lemma 7, any model of $F \wedge G$ satisfies $M$. Since $M$ is inconsistent we conclude $F \wedge G$ is unsatisfiable. □

## 6 Review: Abstract DPLL with Learning

Most modern SAT solvers implement such sophisticated techniques as backjumping and learning:

> **Backjumping:** Chronological Backtracking (used in classical DPLL) can be seen as a prototype of Backjumping. Unlike Backtracking that undoes only the previously made decision, Backjumping is generally able to backtrack further in the search tree by undoing several decisions at once.
> **Learning**: Most modern SAT solvers implement so called *conflict-driven backjumping and learning*: whenever backjumping is performed they add (learn) a "backjump clause" to the clause database of a solver. Learning backjump clauses prevents a solver from reaching "similar" inconsistent states.

In this section we will extend the graph $\text{DP}_F$ to capture the ideas behind backjumping and learning. The new graph will be closely related to the *DPLL System with Learning* graph introduced in (Nieuwenhuis et al. 2006, Section 2.4).

We first note that the graph $\text{DP}_F$ is not adequate to capture such technique as learning since it is incapable to reflect a change in a state of computation related to newly learned clauses. We start by redefining a state so that it incorporates information about changes performed on a clause database.

For a CNF formula $F$, an *augmented state* relative to $F$ is either a distinguished state *FailState* or a pair $M||\Gamma$ where $M$ is a record relative to the set of atoms occurring in $F$, and $\Gamma$ is a (multi-)set of clauses over atoms of $F$ that are entailed by $F$.

*Abstract Answer Set Solvers with Backjumping and Learning (long version)* 21*Unit Propagate* $\lambda$:

$M||\Gamma \Longrightarrow M\,l\,||\Gamma$ if $\begin{cases} C \vee l \in F \cup \Gamma \text{ and} \\ \overline{C} \subseteq M \end{cases}$

*Backjump*:

$P\,l^{\Delta}\,Q\,||\Gamma \Longrightarrow P\,l'\,||\Gamma$ if $\begin{cases} P\,l^{\Delta}\,Q \text{ is inconsistent and} \\ F \models l' \vee \overline{P} \end{cases}$

*Learn*:

$M||\Gamma \Longrightarrow M||C,\Gamma$ if $\begin{cases} \text{every atom in } C \text{ occurs in } F \text{ and} \\ F \models C \end{cases}$

Fig. 3. The additional transition rules of the graph $\text{DPL}_F$.

We now define a graph $\text{DPL}_F$ for any CNF formula $F$. Its nodes are the augmented states relative to $F$. The transition rules *Decide* and *Fail* of $\text{DP}_F$ are extended to $\text{DPL}_F$ as follows: $M||\Gamma \Longrightarrow M'||\Gamma$ ($M||\Gamma \Longrightarrow \textit{FailState}$) is an edge in $\text{DPL}_F$ justified by *Decide* (*Fail*) if and only if $M \Longrightarrow M'$ ($M \Longrightarrow \textit{FailState}$) is an edge in $\text{DP}_F$ justified by *Decide* (*Fail*). Figure 3 presents the other transition rules of $\text{DPL}_F$. We refer to the transition rules *Unit Propagate* $\lambda$, *Backjump*, *Decide*, and *Fail* of the graph $\text{DPL}_F$ as *Basic*. We say that a node in the graph is *semi-terminal* if no rule other than *Learn* is applicable to it.

We will omit the word "augmented" before "state" when this is clear from a context.

The graph $\text{DPL}_F$ can be used for deciding the satisfiability of a formula $F$ simply by constructing an arbitrary path from node $\emptyset||\emptyset$ to a semi-terminal node:

*Proposition 7*

For any CNF formula $F$,

(a) every path in $\text{DPL}_F$ contains only finitely many edges justified by Basic transition rules,
(b) for any semi-terminal state $M||\Gamma$ of $\text{DPL}_F$ reachable from $\emptyset||\emptyset$, $M$ is a model of $F$,
(c) *FailState* is reachable from $\emptyset||\emptyset$ in $\text{DPL}_F$ if and only if $F$ is unsatisfiable.

On the one hand, Proposition 7 (a) asserts that if we construct a path from $\emptyset||\emptyset$ so that Basic transition rules periodically appear in it then some semi-terminal state will be eventually reached. On the other hand, Proposition 7 (b) and (c) assert that as soon as a semi-terminal state is reached the problem of deciding whether formula $F$ is satisfiable is solved. The proof of this proposition is similar to the proof of Theorem 2.12 from (Nieuwenhuis et al. 2006).

For instance, let $F$ be the formula

$$a \vee b$$
$$\neg a \vee c.$$

Here is a path in $\text{DPL}_F$:

$$
\begin{aligned}
&\emptyset || \emptyset && \Longrightarrow (\textit{Learn}) \\
&\emptyset || b \vee c && \Longrightarrow (\textit{Decide}) \\
&\neg b^\Delta || b \vee c && \Longrightarrow (\textit{Unit Propagate } \lambda) \\
&\neg b^\Delta \, c || b \vee c && \Longrightarrow (\textit{Unit Propagate } \lambda) \\
&\neg b^\Delta \, c \, a || b \vee c &&
\end{aligned}
\tag{17}
$$

Since the state $\neg b^\Delta \, c \, a$ is semi-terminal, Proposition 7 (b) asserts that $\{\neg b, c, a\}$ is a model of $F$.

Recall that the transition rule *Backtrack* of the graph $\text{DP}_F$ – a prototype of *Backjump* – is applicable in any inconsistent state with a decision literal in $\text{DP}_F$. The transition rule *Backjump*, on the other hand, is applicable in any inconsistent state with a decision literal that is reachable from $\emptyset || \emptyset$ (the proof of this statement is similar to the proof of Lemma 2.8 from (Nieuwenhuis et al. 2006)). The application of *Backjump* where $l^\Delta$ is the last decision literal and $l'$ is $\overline{l}$ can be seen as an application of *Backtrack*. This fact shows that *Backjump* is essentially a generalization of *Backtrack*. The subgraph of $\text{DP}_F$ induced by the nodes reachable from $\emptyset$ is basically a subgraph of $\text{DPL}_F$.

## 7 Answer Set Solver with Learning

In this section we will extend the graph $\text{SM}_\Pi$ to capture backjumping and learning. As a result we will be able to model the algorithms of systems $\text{SMODELS}_{cc}$ and SUP.

### 7.1 Graph $\text{SML}_\Pi$

An *(augmented) state* relative to a program $\Pi$ is either a distinguished state *FailState* or a pair of the form $M||\Gamma$ where $M$ is a record relative to the set of atoms occurring in $\Pi$, and $\Gamma$ is a (multi-)set of constraints formed from atoms occurring in $\Pi$ that are entailed by $\Pi$.

For any program $\Pi$, we will define a graph $\text{SML}_\Pi$. Its nodes are the augmented states relative to $\Pi$. The transition rules *Unit Propagate LP*, *All Rules Cancelled*, *Backchain True*, *Unfounded*, *Decide* and *Fail* of $\text{SM}_\Pi$ are extended to $\text{SML}_\Pi$ as follows: $M||\Gamma \Longrightarrow M'||\Gamma$ ($M||\Gamma \Longrightarrow \textit{FailState}$) is an edge in $\text{SML}_\Pi$ justified by a transition rule $T$ if and only if $M \Longrightarrow M'$ ($M \Longrightarrow \textit{FailState}$) is an edge in $\text{SM}_\Pi$ justified by $T$. Figure 4 presents the other transition rules of $\text{SML}_\Pi$.

We refer to the transition rules *Unit Propagate LP*, *All Rules Cancelled*, *Backchain True*, *Backchain False* $\lambda$, *Unfounded*, *Backjump LP*, *Decide*, and *Fail* of the graph $\text{SML}_\Pi$ as *Basic*. We say that a node in the graph is *semi-terminal* if no rule other than *Learn LP* is applicable to it.

The graph $\text{SML}_\Pi$ can be used for deciding whether a program $\Pi$ has an answer set by constructing a path from $\emptyset||\emptyset$ to a semi-terminal node:

*Proposition 8*
For any program $\Pi$,

*Backchain False* $\lambda$:

$$M||\Gamma \Longrightarrow M\,\overline{l}||\Gamma \ \text{ if } \begin{cases} a \leftarrow l, B \in \Pi \cup \Gamma, \\ \neg a \in M \text{ or } a = \bot, \\ B \subseteq M \end{cases}$$

*Backjump LP*:

$$P\,l^{\Delta}\,Q||\Gamma \Longrightarrow P\,l'||\Gamma \ \text{ if } \begin{cases} P\,l^{\Delta}\,Q \text{ is inconsistent and} \\ \Pi \text{ entails } l' \vee \overline{P} \end{cases}$$

*Learn LP*:

$M||\Gamma \Longrightarrow M|| \leftarrow B, \Gamma \ \text{ if } \ \Pi \text{ entails } \overline{B}$

Fig. 4. The additional transition rules of the graph $\text{SML}_\Pi$.

(a) every path in $\text{SML}_\Pi$ contains only finitely many edges labeled by Basic transition rules,
(b) for any semi-terminal state $M||\Gamma$ of $\text{SML}_\Pi$ reachable from $\emptyset||\emptyset$, $M^+$ is an answer set of $\Pi$,
(c) *FailState* is reachable from $\emptyset||\emptyset$ in $\text{SML}_\Pi$ if and only if $\Pi$ has no answer sets.

Thus if we construct a path from $\emptyset||\emptyset$ so that Basic transition rules periodically appear in it then some semi-terminal state will be eventually reached; as soon as a semi-terminal state is reached the problem of finding an answer set is solved.

For instance, let $\Pi$ be program (8). Here is a path in $\text{SML}_\Pi$ with every edge annotated by the name of a transition rule that justifies the presence of this edge in the graph :

$$\begin{array}{lll}
\emptyset||\emptyset & \Longrightarrow & (Decide) \\
a^{\Delta}||\emptyset & \Longrightarrow & (Unit\ Propagate\ LP) \\
a^{\Delta}\,c||\emptyset & \Longrightarrow & (All\ Rules\ Cancelled) \\
a^{\Delta}\,c\,\neg b||\emptyset & \Longrightarrow & (Decide) \\
a^{\Delta}\,c\,\neg b\,d^{\Delta}||\emptyset & \Longrightarrow & (Unfounded) \\
a^{\Delta}\,c\,\neg b\,d^{\Delta}\,\neg d||\emptyset & \Longrightarrow & (Backjump\ LP) \\
a^{\Delta}\,c\,\neg b\,\neg d||\emptyset & \Longrightarrow & (Learn\ LP) \\
a^{\Delta}\,c\,\neg b\,\neg d||\neg a \vee \neg c \vee b \vee \neg d &&
\end{array} \qquad (18)$$

Since the state $a^{\Delta}\,c\,\neg b\,\neg d$ is semi-terminal, Proposition 8 (b) asserts that

$$\{a, c, \neg b, \neg d\}^+ = \{a, c\}$$

is an answer set for $\Pi$.

Proof of Proposition 8 is in Section 10.

As in case of the graphs $\text{DP}_F$ and $\text{DPL}_F$, *Backjump LP* is applicable in any inconsistent state with a decision literal that is reachable from $\emptyset||\emptyset$ (Proposition 11 from Section 9), and is essentially a generalization of the transition rule *Backtrack* of the graph $\text{SM}_\Pi$.

Modern SAT solvers often implement such sophisticated techniques as restart and forgetting in addition to backjumping and learning:

**Restart**: A solver restarts the DPLL procedure whenever the search is not making "enough" progress. The idea is that upon a restart a solver will explore a new part of the search space using the clauses that have been learned.

**Forgetting**: This technique is usually implemented in relation with conflict-driven backjumping and learning. When a solver "notes" that earlier learned clauses are not helpful anymore it removes (forgets) them from the clause database. Forgetting allows a solver to avoid a possible exponential space blow-up introduced by learning.

We may extend the graph $\text{sml}_\Pi$ with the following transition rules that capture the ideas behind these technique:

$$\textit{Restart}:$$
$$M||\Gamma \Longrightarrow \emptyset||\Gamma$$

$$\textit{Forget LP}:$$
$$M|| \leftarrow B, \Gamma \Longrightarrow M||\Gamma.$$

The transition rules *Restart* and *Forget LP* are similar to the analogous rules in (Nieuwenhuis et al. 2006) for extending DPLL procedure with restart and forgetting techniques. It is easy to prove a result similar to Proposition 8 for the graph $\text{sml}_\Pi$ with *Restart* and *Forget LP* (for such graph a state is semi-terminal if no rule other than *Learn LP*, *Restart*, *Forget LP* is applicable to it.)

### 7.2 Smodels$_{cc}$ and Sup Algorithms

In Section 4.2 we demonstrated a method for specifying the algorithm of an answer set solver by means of the graph $\text{sm}_\Pi$. In particular, we described the SMODELS algorithm by assigning priorities to transition rules of $\text{sm}_\Pi$. In this section we use this method to describe the SMODELS$_{cc}$ (Ward and Schlipf 2004) and SUP (Lierler 2008) algorithms by means of $\text{sml}_\Pi$.

System SMODELS$_{cc}$ enhances the SMODELS algorithm with conflict-driven backjumping and learning. Its strategy for choosing a path in the graph $\text{sml}_\Pi$ is similar to that of SMODELS. System SMODELS$_{cc}$ assigns priorities to inference rules of $\text{sml}_\Pi$ as follows:

*Backjump LP*, *Fail* $\gg$
*Unit Propagate LP*, *All Rules Cancelled*, *Backchain True*, *Backchain False* $\lambda \gg$
*Unfounded* $\gg$
*Decide*.

Also, SMODELS$_{cc}$ always applies the transition rule *Learn LP* in a non-semi-terminal state reached by an application of *Backjump LP*, because it implements conflict-driven backjumping and learning.[12] In Section 11 we discuss details on which clause is being learned during the application of *Learn LP*.

---

[12] System SMODELS$_{cc}$ (SUP) also implements restarts and forgetting that may be modeled by the transition rules *Restart* and *Forget LP*. An application of these transition rules in $\text{sml}_\Pi$ relies on particular heuristics implemented by the solver.

In (Lierler 2008), we introduced the simplified SUP algorithm that relies on backtracking rather than conflict-driven backjumping and learning that are actually implemented in the system. We now present the SUP algorithm that takes these sophisticated techniques into account.

System SUP assigns priorities to inference rules of SML$_\Pi$ as follows:

*Backjump LP, Fail* $\gg$
*Unit Propagate LP, All Rules Cancelled, Backchain True, Backchain False* $\lambda \gg$
*Decide* $\gg$
*Unfounded.*

Similarly to SMODELS$_{cc}$, SUP always applies the transition rule *Learn LP* in a non-semi-terminal state reached by an application of *Backjump LP*.

For example, let $\Pi$ be program (8). Path (18) corresponds to an execution of system SUP, but does not correspond to any execution of SMODELS$_{cc}$ because for the latter *Unfounded* is a rule of higher priority than *Decide*. Here is another path in SML$_\Pi$ from $\emptyset||\emptyset$ to the same semi-terminal node:

$$
\begin{array}{ll}
\emptyset||\emptyset & \Longrightarrow (\textit{Decide}) \\
a^\Delta||\emptyset & \Longrightarrow (\textit{Unit Propagate LP}) \\
a^\Delta c||\emptyset & \Longrightarrow (\textit{All Rules Cancelled}) \\
a^\Delta c \neg b||\emptyset & \Longrightarrow (\textit{Unfounded}) \\
a^\Delta c \neg b \neg d||\emptyset &
\end{array}
\quad (19)
$$

Path (19) corresponds to an execution of system SMODELS$_{cc}$, but does not correspond to any execution of system SUP because for the latter *Decide* is a rule of higher priority than *Unfounded*.

The strategy of SUP of assigning the transition rule *Unfounded* the lowest priority may be reasonable for many problems. For instance, it is easy to see that transition rule *Unfounded* is redundant for tight programs. The SUP algorithm is similar to SAT-based answer set solvers such as ASSAT (Lin and Zhao 2004) and CMODELS (Giunchiglia et al. 2006) (see Section 8.2) in the fact that it will first compute a supported model of a program and only then will test whether this model is indeed an answer set, i.e., whether *Unfounded* is applicable in this state.

## 8 Generate and Test with Learning

In this section we model backjumping and learning for the *generate and test* procedure by defining a graph GTL$_{F,G}$ that extends GT$_{F,G}$ (Section 5) in a similar manner as DPL$_F$ (Section 6) extends DP$_F$.

### 8.1 Graph GTL$_{F,G}$

An *(augmented) state* relative to a CNF formula $F$ and a formula $G$ formed from atoms occurring in $F$ is either a distinguished state *FailState* or a pair of the form $M||\Gamma$, where $M$ is a record (Section 2.1) relative to the set of atoms occurring in $F$, and $\Gamma$ is a (multi-)set of clauses formed from atoms occurring in $F$ that are entailed by $F \wedge G$.

The nodes of the graph $\text{GTL}_{F,G}$ are the augmented states relative to a CNF formula $F$ and a formula $G$ formed from atoms occurring in $F$. The edges of $\text{GTL}_{F,G}$ are described by the transition rules *Unit Propagate* $\lambda$, *Decide*, *Fail* of $\text{DPL}_F$, the transition rules

*Backjump GT* :
$$P\,l^\Delta\,Q||\Gamma \Longrightarrow P\,l'||\Gamma \ \ \text{if} \ \begin{cases} P\,l^\Delta\,Q \text{ is inconsistent and} \\ F \wedge G \models l' \vee \overline{P} \end{cases}$$

*Learn GT*:
$$M||\Gamma \Longrightarrow M||C,\Gamma \ \ \text{if} \ \begin{cases} \text{every atom in } C \text{ occurs in } F \text{ and} \\ F \wedge G \models C \end{cases}$$

and the transition rule *Test* of $\text{GT}_{F,G}$ that is extended to $\text{GTL}_{F,G}$ as follows: $M||\Gamma \Longrightarrow M'||\Gamma$ is an edge in $\text{GTL}_{F,G}$ justified by *Test* if and only if $M \Longrightarrow M'$ is an edge in $\text{GT}_{F,G}$ justified by *Test*.

We refer to the transition rules *Unit Propagate* $\lambda$, *Test*, *Decide*, *Fail*, *Backjump GT* of the graph $\text{GTL}_{F,G}$ as *Basic*. We say that a node in the graph is *semi-terminal* if no rule other than *Learn GT* is applicable to it.

The graph $\text{GTL}_{F,G}$ can be used for deciding whether a formula $F \wedge G$ has a model by constructing a path from $\emptyset||\emptyset$ to a terminal node:

*Proposition 9*
For any CNF formula $F$ and a formula $G$ formed from atoms occurring in $F$,

(a) every path in $\text{GTL}_{F,G}$ contains only finitely many edges labeled by Basic transition rules,
(b) for any semi-terminal state $M||\Gamma$ of $\text{GTL}_{F,G}$ reachable from $\emptyset||\emptyset$, $M$ is a model of $F \wedge G$,
(c) *FailState* is reachable from $\emptyset||\emptyset$ in $\text{GTL}_{F,G}$ if and only if $F \wedge G$ is unsatisfiable.

As in case of the graph $\text{DPL}_F$, the transition rule *Backjump GT* is applicable in any inconsistent state with a decision literal that is reachable from $\emptyset||\emptyset$. We call such states *backjump* states.

*Proposition 10*
For any CNF formula $F$ and a formula $G$ formed from atoms occurring in $F$, the transition rule *Backjump GT* is applicable in any backjump state in $\text{GTL}_{F,G}$.

Proofs of Propositions 9 and 10 are given in Section 13.

### 8.2 Cmodels Algorithm

System CMODELS implements an algorithm called ASP-SAT with Learning (Giunchiglia et al. 2006) that extends ASP-SAT with Backtracking by backjumping and learning.

The application of CMODELS to a program $\Pi$ can be viewed as constructing a path from $\emptyset||\emptyset$ to a terminal node in the graph $\text{GTL}_{F,G}$, where

- $F$ is the completion of $\Pi$ converted to conjunctive normal form, and
- $G$ is $LF(\Pi)$.

In Sections 4.2 we demonstrated a method for specifying the algorithm of an answer set solver by means of the graph SM$_\Pi$. We use this method to describe the CMODELS algorithm using the graph GTL$_{F,G}$. System CMODELS assigns priorities to the inference rules of GTL$_{F,G}$ as follows:

$$\begin{aligned}&\textit{Backjump GT, Fail} \gg \\ &\textit{Unit Propagate } \lambda \gg \\ &\textit{Decide} \gg \\ &\textit{Test}.\end{aligned}$$

Also, CMODELS always applies the transition rule *Learn GT* in a non-semi-terminal state reached by an application of *Backjump GT*.

The priorities imposed on the rules by CMODELS guarantee that the transition rule *Test* is applied to a model of $F \cup \Gamma$ (clausified completion $F$ extended by learned clauses $\Gamma$). This allows CMODELS to proceed with its search in case if a found model is not an answer set. Furthermore, the CMODELS strategy guarantees that in a state reached by an application of *Test*, first *Backjump GT* will be applied and then in the resulting state *Learn GT* will be applied. The clause learned due to this application of *Learn GT* is derived by means of loop formulas (see (Giunchiglia et al. 2006)). In this sense CMODELS uses loop formulas to guide its search.

Systems SAG (Lin et al. 2006) and CLASP (Gebser et al. 2007) are answer set solvers that are enhancements of CMODELS. First, they compute and clausify program's completion and then use unit propagate on resulting propositional formula as an inference mechanism. Second, they guide their search by means of loop formulas. Third, they implement conflict-driven backjumping and learning. Also, SAG uses SAT solvers for search. The systems differ from CMODELS in the following:

- they maintain the data structure representing an input logic program through out the whole computation,
- in addition to implementing inference rules of the graph GTL$_{F,G}$ they also implement the inference rule *Unfounded* of SM$_\Pi$. A hybrid graph combining the inference rule *Unfounded* of SM$_\Pi$ and the inference rules of GTL$_{F,G}$ may be used to describe the SAG and CLASP algorithms.

System SAG assigns the same priorities to the inference rules of the hybrid graph as CMODELS. Also, SAG at random decides whether to apply the inference rule *Unfounded* in a state.

On the other hand, system CLASP assigns priorities to the inference rules of the hybrid graph as follows:

$$\begin{aligned}&\textit{Backjump GT, Fail} \gg \\ &\textit{Unit Propagate } \lambda, \textit{Unfounded} \gg \\ &\textit{Decide}.\end{aligned}$$

Like CMODELS, both SAG and CLASP always apply the transition rule *Learn GT* in a non-semi-terminal state reached by an application of *Backjump GT*.

## 9 Backjumping and Extended Graph

Recall the transition rule *Backjump LP* of $\text{SML}_\Pi$

*Backjump LP*:
$$P l^\Delta Q || \Gamma \Longrightarrow P l' || \Gamma \quad \text{if} \quad \begin{cases} P l^\Delta Q \text{ is inconsistent and} \\ \Pi \text{ entails } l' \vee \overline{P}. \end{cases}$$

A state in the graph $\text{SML}_\Pi$ is a *backjump state* if it is inconsistent, contains a decision literal, and is reachable from $\emptyset || \emptyset$. Note that it may be not clear a priori whether *Backjump LP* is applicable to a backjump state and if so to which state the edge due to the application of *Backjump LP* leads. These questions are important if we want to base an algorithm on this framework. It turns out that *Backjump LP* is always applicable to a backjump state:

*Proposition 11*
For a program $\Pi$, the transition rule *Backjump LP* is applicable to any backjump state in $\text{SML}_\Pi$.

Proposition 11 guarantees that a backjump state in $\text{SML}_\Pi$ is never semi-terminal. In the end of this section we show how Proposition 11 can be derived from the results proved later in this paper. Next question to answer is how to continue choosing a path in the graph after reaching a backjump state. To answer this question we introduce the notions of reason and extended graph.

For a program $\Pi$, we say that a clause $l \vee C$ is a *reason* for $l$ to be in a list of literals $P l Q$ w.r.t $\Pi$ if $\Pi$ entails $l \vee C$ and $\overline{C} \subseteq P$. We can equivalently restate the second condition of *Backjump LP* "$\Pi$ entails $l' \vee \overline{P}$" as "there exists a reason for $l'$ to be in $P l'$ w.r.t. $\Pi$" (note that $l' \vee \overline{P}$ is a reason for $l'$ to be in $P l'$). We call a reason for $l'$ to be in $P l'$ a *backjump clause*. Note that Proposition 11 asserts that a backjump clause always exists for a backjump state. It is clear that we may continue choosing a path in the graph after reaching a backjump state if we know how to compute a backjump clause for this state. We now define a graph $\text{SML}_\Pi^\uparrow$ that shares many properties of $\text{SML}_\Pi$ but allows us to give a simpler procedure for computing a backjump clause.

An *extended record* $M$ relative to a program $\Pi$ is a list of literals over the set of atoms occurring in $\Pi$ where

(i) each literal $l$ in $M$ is annotated either by $\Delta$ or by a reason for $l$ to be in $M$ w.r.t. $\Pi$,
(ii) $M$ contains no repetitions,
(iii) for any inconsistent prefix of $M$ its last literal is annotated by a reason.

For instance, let $\Pi$ be the program

$$a \leftarrow \; not \; b$$
$$c.$$

The list of literals

$$b^\Delta \, a^\Delta \, \neg b^{\neg b \vee \neg a}$$

is an extended record relative to $\Pi$. On the other hand, the lists of literals

$$a^{\Delta}\neg a^{\Delta} \qquad a^{\Delta}\neg b^{\neg b\vee\neg a}\, b^{\Delta} \qquad b^{\Delta}\, a^{\Delta}\neg b^{\neg b\vee\neg a}\, c^{\Delta}$$

are not extended records.

An *extended state* relative to a program $\Pi$ is either a distinguished state *FailState* or a pair of the form $M||\Gamma$ where $M$ is an extended record relative to $\Pi$, and $\Gamma$ is the same as in the definition of an augmented state (i.e., $\Gamma$ is a (multi-)set of constraints formed from atoms occurring in $\Pi$ that are entailed by $\Pi$.) It is easy to see that for any extended state $S$ relative to a program $\Pi$, the result of removing annotations from all nondecision literals of $S$ is a state of $\text{SML}_\Pi$: we will denote this state by $S^{\downarrow}$.

For instance, consider program $a \leftarrow \ not\ b$. All pairs

$$\textit{FailState} \quad \emptyset||\emptyset \quad a^{\Delta}\neg b^{\neg b\vee\neg a}||\emptyset \quad \neg a^{\Delta}\, b^{b\vee a}||\emptyset$$

are among valid extended states relative to this program. The corresponding states $S^{\downarrow}$ are

$$\textit{FailState} \quad \emptyset||\emptyset \quad a^{\Delta}\neg b||\emptyset \quad \neg a^{\Delta}\, b||\emptyset.$$

We now define a graph $\text{SML}_\Pi^{\uparrow}$ for any program $\Pi$. Its nodes are the extended states relative to $\Pi$. The transition rules of $\text{SML}_\Pi$ are extended to $\text{SML}_\Pi^{\uparrow}$ as follows: $S_1 \Longrightarrow S_2$ is an edge in $\text{SML}_\Pi^{\uparrow}$ justified by a transition rule $T$ if and only if $S_1^{\downarrow} \Longrightarrow S_2^{\downarrow}$ is an edge in $\text{SML}_\Pi$ justified by $T$.

We will omit the word "extended" before "record" and "state" when this is clear from a context.

The following lemma formally states the relationship between nodes of the graphs $\text{SML}_\Pi$ and $\text{SML}_\Pi^{\uparrow}$:

*Lemma 8*
For any program $\Pi$, if $S'$ is a state reachable from $\emptyset||\emptyset$ in the graph $\text{SML}_\Pi$ then there is a state $S$ in the graph $\text{SML}_\Pi^{\uparrow}$ such that $S^{\downarrow} = S'$.

The definitions of Basic transition rules and semi-terminal states in $\text{SML}_\Pi^{\uparrow}$ are similar to their definitions for $\text{SML}_\Pi$.

*Proposition $8^{\uparrow}$*
For any program $\Pi$,

(a) every path in $\text{SML}_\Pi^{\uparrow}$ contains only finitely many edges labeled by Basic transition rules,
(b) for any semi-terminal state $M||\Gamma$ of $\text{SML}_\Pi^{\uparrow}$, $M^+$ is an answer set of $\Pi$,
(c) $\text{SML}_\Pi^{\uparrow}$ contains an edge leading to *FailState* if and only if $\Pi$ has no answer sets.

Note that Proposition $8^{\uparrow}$ (b), unlike Proposition 8 (b), is not limited to semi-terminal states that are reachable from $\emptyset||\emptyset$. As in the case of the graph $\text{SML}_\Pi$, $\text{SML}_\Pi^{\uparrow}$ can be used for deciding whether a program $\Pi$ has an answer set. Furthermore, the new graph provides the means for computing a backjump clause that permits practical application of the transition rule *Backjump LP*: Sections 10.3 and 11 describe the *BackjumpClause* (Algorithm 1) and *BackjumpClauseFirstUIP* (Algorithm 2) procedures that compute *Decision* and *FirstUIP* backjump clauses respectively.

We say that a state in the graph $\text{SML}_\Pi^\uparrow$ is a *backjump state* if its record is inconsistent and contains a decision literal. Unlike the definition of a backjump state in $\text{SML}_\Pi$, this definition does not require a backjump state to be reachable from $\emptyset || \emptyset$ in $\text{SML}_\Pi^\uparrow$. As in case of the graph $\text{SML}_\Pi$, any backjump state in $\text{SML}_\Pi^\uparrow$ is not semi-terminal:

*Proposition 11$^\uparrow$*

For a program $\Pi$, the transition rule *Backjump LP* is applicable to any backjump state in $\text{SML}_\Pi^\uparrow$.

Proposition 8 (b), (c) and Proposition 11 easily follow from Lemma 8 and Proposition 8$^\uparrow$ (b), (c) and Proposition 11$^\uparrow$ respectively. Proof of Proposition 8 (a) is similar to the proof of Proposition 8$^\uparrow$ (a).

Next section will present the proofs for Proposition 8$^\uparrow$, Lemma 8, and Proposition 11$^\uparrow$. It is interesting to note that the proofs of Lemma 8 and Proposition 11$^\uparrow$ implicitly provide the means for choosing a path in the graph $\text{SML}_\Pi^\uparrow$:

- given a state $M||\Gamma$ and a transition rule *Unit Propagate LP*, *All Rules Cancelled*, *Backchain True*, *Backchain False* $\lambda$, or *Unfounded* applicable to $M||\Gamma$, the proof of Lemma 8 describes a clause that may be used to construct a record $M'$ so that there is an edge $M||\Gamma \implies M'||\Gamma$ due to this transition rule,
- given a backjump state $M||\Gamma$, the proof of Proposition 11$^\uparrow$ describes a backjump clause that can be used to construct a record $M'$ so that there is an edge $M||\Gamma \implies M'||\Gamma$ due to *Backjump LP*.

Furthermore, the construction of the proof of Proposition 11$^\uparrow$ paves the way for procedure *BackjumpClause* presented in Algorithm 1.

## 10  Proofs of Proposition 8$^\uparrow$, Lemma 8, Proposition 11$^\uparrow$

### 10.1  Proof of Proposition 8$^\uparrow$

*Lemma 9*

For any program $\Pi$, an extended record $M$ relative to $\Pi$, and every assignment $X$ such that $X^+$ is an answer set for $\Pi$, if $X$ satisfies all decision literals in $M$ then $X \models M$.

*Proof*

By induction on the length of $M$. The property trivially holds for $\emptyset$. We assume that the property holds for any state with $n$ elements. Consider any state $M$ with $n+1$ elements. Let $X$ be an assignment such that $X^+$ is an answer set for $\Pi$ and $X$ satisfies all decision literals in $M$. We will now show that $X \models M$.

Case 1. $M$ has the form $P\,l^\Delta$. By the inductive hypothesis, $X \models P$. Since $X$ satisfies all decision literals in $M$, $X \models l$.

Case 2. $M$ has the form $P\,l^{l\vee C}$. By the inductive hypothesis, $X \models P$. By the definition of a reason, (i) $\Pi$ entails $l \vee C$ and (ii) $\overline{C} \subseteq P$. From (ii) it follows that $P \models \neg C$. Consequently, $X \models \neg C$. From (i) it follows that for any assignment $X$ such that $X^+$ is an answer set, $X \models l \vee C$. Consequently, $X \models l$. □

The proof of Proposition $8^\uparrow$ assumes the correctness of Proposition $11^\uparrow$ that we demonstrate in Section 10.3.

*Proposition $8^\uparrow$*
For any program $\Pi$,

(a) every path in $\text{SML}_\Pi^\uparrow$ contains only finitely many edges labeled by Basic transition rules,
(b) for any semi-terminal state $M||\Gamma$ of $\text{SML}_\Pi^\uparrow$, $M^+$ is an answer set of $\Pi$,
(c) $\text{SML}_\Pi^\uparrow$ contains an edge leading to *FailState* if and only if $\Pi$ has no answer sets.

*Proof*
(a) For any list $N$ of literals by $|N|$ we denote the length of $N$. Any state $M||\Gamma$ has the form $M_0\,l_1^\Delta\,M_1\ldots l_p^\Delta\,M_p||\Gamma$, where $l_1^\Delta\ldots l_p^\Delta$ are all decision literals of $M$; we define $\alpha(M||\Gamma)$ as the sequence of nonnegative integers $|M_0|,|M_1|,\ldots,|M_p|$, and $\alpha(\textit{FailState}) = \infty$. For any states $S$ and $S'$ of $\text{SML}_\Pi^\uparrow$, we understand $\alpha(S) < \alpha(S')$ as the lexicographical order. We first note that for any state $M||\Gamma$, value of $\alpha$ is based only on the first component $M$ of the state. Second, there is a finite number of distinct values of $\alpha$ due to the fact that there is a finite number of distinct $M$s over $\Pi$. We conclude that there is a finite number of distinct values of $\alpha$ for the states of $\text{SML}_\Pi^\uparrow$, even though the number of distinct states in $\text{SML}_\Pi^\uparrow$ is infinite.

By the definition of the transition rules of $\text{SML}_\Pi^\uparrow$, if there is an edge from $M||\Gamma$ to $M'||\Gamma'$ in $\text{SML}_\Pi^\uparrow$ formed by any Basic transition rule then $\alpha(M||\Gamma) < \alpha(M'||\Gamma')$. Then, due to the fact that there is a finite number of distinct values of $\alpha$, it follows that there is only a finite number of edges due to the application of Basic rules possible in any path.

(b) Let $M||\Gamma$ be a semi-terminal state so that none of the Basic rules are applicable. From the fact that *Decide* is not applicable, we conclude that $M$ assigns all literals.

Furthermore, $M$ is consistent. Indeed, assume that $M$ is inconsistent. Then, since *Fail* is not applicable, $M$ contains a decision literal. Consequently, $M||\Gamma$ is a backjump state. By Proposition $11^\uparrow$, the transition rule *Backjump LP* is applicable in $M||\Gamma$. This contradicts our assumption that $M||\Gamma$ is semi-terminal.

Also, $M$ is a model of $\Pi$: since *Unit Propagate LP* is not applicable in $M||\Gamma$, it follows that for every rule $a \leftarrow B \in \Pi$, if $B \subseteq M$ then $a \in M$.

Assume that $M^+$ is not an answer set. Then, by Lemma 3, there is a non-empty unfounded set $U$ on $M$ w.r.t. $\Pi$ such that $U \subseteq M$. It follows that *Unfounded* is applicable (with an arbitrary $a \in U$) in $M||\Gamma$. This contradicts the assumption that $M||\Gamma$ is semi-terminal.

(c) Left-to-right: There is a state $M||\Gamma$ in $\text{SML}_\Pi^\uparrow$ such that there is an edge between $M||\Gamma$ and *FailState*. By the definition of $\text{SML}_\Pi^\uparrow$, this edge is due to the transition rule *Fail*. Consequently, state $M||\Gamma$ is such that $M$ is inconsistent and contains

no decision literals. By Lemma 9, for every assignment $X$ such that $X^+$ is an answer set for $\Pi$, $X$ satisfies $M$. Since $M$ is inconsistent we conclude that $\Pi$ has no answer sets.

Right-to-left: Consider the process of constructing a path consisting only of edges due to Basic transition rules. By (a), it follows that this path will eventually reach a semi-terminal state. By (b), this semi-terminal state cannot be different from *FailState*, because $\Pi$ has no answer sets. We conclude that there is an edge leading to *FailState*.  □

### *10.2 Proof of Lemma 8*

The proof uses the notion of loop formula (Lin and Zhao 2004).

Given a set $A$ of atoms by $Bodies(\Pi, A)$ we denote the set that consists of the elements of $Bodies(\Pi, a)$ for all $a$ in $A$. Let $\Pi$ be a program. For any set $Y$ of atoms, the *external support formula* (Lee 2005) for $Y$ is

$$\bigvee_{B \in Bodies(\Pi, Y), B^+ \cap Y = \emptyset} B. \tag{20}$$

We will denote the external support formula by $ES_{\Pi, Y}$. For any set $Y$ of atoms, the *loop formula* for $Y$ is the implication

$$\bigvee_{a \in Y} a \to ES_{\Pi, Y}.$$

We can rewrite this formula as the disjunction

$$\bigwedge_{a \in Y} \neg a \vee ES_{\Pi, Y}. \tag{21}$$

From the *Main Theorem* in (Lee 2005) we conclude:

*Lemma on Loop Formulas*
For any program $\Pi$, $\Pi$ entails loop formulas (21) for all sets $Y$ of atoms that occur in $\Pi$.

For a state $S$ in the graph $\text{SML}_\Pi^\uparrow$, we say that $S^\downarrow$ in $\text{SML}_\Pi$ is the *image* of $S$.

*Lemma 8*
For any program $\Pi$, if $S'$ is a state reachable from $\emptyset || \emptyset$ in the graph $\text{SML}_\Pi$ then there is a state $S$ in the graph $\text{SML}_\Pi^\uparrow$ such that $S^\downarrow = S'$.

*Proof*
Since the property trivially holds for the initial state $\emptyset || \emptyset$, we only need to prove that all transition rules of $\text{SML}_\Pi$ preserve it.

Consider an edge $M || \Gamma \Longrightarrow M' || \Gamma'$ in the graph $\text{SML}_\Pi$ such that there is a state $M_1 || \Gamma$ in the graph $\text{SML}_\Pi^\uparrow$ satisfying the condition $(M_1 || \Gamma)^\downarrow = M || \Gamma$. We need to show that there is a state in the graph $\text{SML}_\Pi^\uparrow$ such that $M' || \Gamma'$ is its image in $\text{SML}_\Pi$. Consider several cases that correspond to a transition rule leading from $M || \Gamma$ to $M' || \Gamma'$:

*Unit Propagate LP*:

$$M||\Gamma \Longrightarrow M\ a||\Gamma \ \text{if} \ \begin{cases} a \leftarrow B \in \Pi \ \text{and} \\ B \subseteq M. \end{cases}$$

$M'||\Gamma'$ is $M\,a||\Gamma$. It is sufficient to prove that $M_1\,a^{a \vee \overline{B}}||\Gamma$ is a state of $\text{SML}_\Pi^\uparrow$. It is enough to show that a clause $a \vee \overline{B}$ is a reason for $a$ to be in $M\,a$. By applicability conditions of *Unit Propagate LP*, $B \subseteq M$. Since $\Pi$ entails its rule $a \leftarrow B$, $\Pi$ entails $a \vee \overline{B}$.

*All Rules Cancelled*:

$$M||\Gamma \Longrightarrow M\ \neg a||\Gamma \ \text{if} \ \overline{B} \cap M \neq \emptyset \ \text{for all} \ B \in Bodies(\Pi, a).$$

$M'||\Gamma'$ is $M \neg a||\Gamma$. Consider any $B \in Bodies(\Pi, a)$. Since $\overline{B} \cap M \neq \emptyset$, $B$ contains a literal from $\overline{M}$: call it $f(B)$. It is sufficient to show that

$$\neg a \vee \bigvee_{B \in Bodies(\Pi, a)} f(B) \tag{22}$$

is a reason for $\neg a$ to be in $M \neg a$.

First, by the choice of $f(B)$, $f(B) \in \overline{M}$; consequently,

$$\overline{\bigvee_{B \in Bodies(\Pi, a)} f(B)} \subseteq M.$$

Second, since $f(B) \in B$, the loop formula $\neg a \vee ES_{\Pi, \{a\}}$ entails (22). By *Lemma on Loop Formulas*, it follows that $\Pi$ entails (22).

*Backchain True*:

$$M||\Gamma \Longrightarrow M\,l||\Gamma \ \text{if} \ \begin{cases} a \leftarrow B \in \Pi, \\ a \in M, \\ \overline{B'} \cap M \neq \emptyset \ \text{for all} \ B' \in Bodies(\Pi, a) \setminus \{B\}, \\ l \in B. \end{cases}$$

$M'||\Gamma'$ is $M\,l||\Gamma$. Consider any $B' \in Bodies(\Pi, a) \setminus B$. Since $\overline{B'} \cap M \neq \emptyset$, $B'$ contains a literal from $\overline{M}$: call it $f(B')$. A clause

$$l \vee \neg a \vee \bigvee_{B' \in Bodies(\Pi, a) \setminus B} f(B') \cdot \tag{23}$$

is a reason for $l$ to be in $M\,l$. The proof of this statement is similar to the case of *All Rules Cancelled*.

*Backchain False $\lambda$*:

$$M||\Gamma \Longrightarrow M\ \overline{l}||\Gamma \ \text{if} \ \begin{cases} a \leftarrow l, B \ \in \ \Pi \cup \Gamma, \\ \neg a \in M \ \text{or} \ a = \bot, \\ B \subseteq M. \end{cases}$$

$M'||\Gamma'$ is $M\,\overline{l}||\Gamma$. A clause $\overline{l} \vee \overline{B} \vee a$ is a reason for $\overline{l}$ to be in $M\overline{l}$. The proof of this statement is similar to the case of *Unit Propagate LP*.

*Unfounded*:

$$M||\Gamma \Longrightarrow M \neg a||\Gamma \ \text{if} \ \begin{cases} M \ \text{is consistent and} \\ a \in U \ \text{for a set} \ U \ \text{unfounded on} \ M \ \text{w.r.t.} \ \Pi. \end{cases}$$

$M'||\Gamma'$ is $M \neg a||\Gamma$. Consider any $B \in Bodies(\Pi, U)$ such that $U \cap B^+ = \emptyset$. By the definition of an unfounded set, it follows that $\overline{B} \cap M \neq \emptyset$. Consequently, $B$ contains a literal from $\overline{M}$: call it $f(B)$. The clause

$$\neg a \vee \bigvee_{Bodies(\Pi, U), B^+ \cap U = \emptyset} f(B) \tag{24}$$

is a reason for $\neg a$ to be in $M \neg a$. The proof of this statement is similar to the case of *All Rules Cancelled*.

*Backjump LP*, *Decide*, *Fail*, and *Learn LP*: obvious. $\square$

The process of turning a state of $\text{SML}_\Pi$ reachable from $\emptyset||\emptyset$ into a corresponding state of $\text{SML}_\Pi^\uparrow$ can be illustrated by the following example: Consider a program $\Pi$

$$\begin{aligned}
&a \leftarrow not\ b \\
&b \leftarrow not\ a,\ not\ c \\
&c \leftarrow not\ f \\
&\leftarrow k,\ d \\
&k \leftarrow l,\ not\ b \\
&\leftarrow m,\ not\ l,\ not\ b \\
&m \leftarrow\ not\ k,\ not\ l
\end{aligned} \tag{25}$$

and a path in $\text{SML}_\Pi$

$$\begin{aligned}
&\emptyset||\emptyset \Longrightarrow (Decide) \\
&a^\Delta||\emptyset \Longrightarrow (All\ Rules\ Cancelled) \\
&a^\Delta \neg b||\emptyset \Longrightarrow (Decide) \\
&a^\Delta \neg b\, c^\Delta||\emptyset \Longrightarrow (Backchain\ True) \\
&a^\Delta \neg b\, c^\Delta \neg f||\emptyset \Longrightarrow (Decide) \\
&a^\Delta \neg b\, c^\Delta \neg f\, d^\Delta||\emptyset \Longrightarrow (Backchain\ False\ \lambda) \\
&a^\Delta \neg b\, c^\Delta \neg f\, d^\Delta \neg k||\emptyset \Longrightarrow (Backchain\ False\ \lambda) \\
&a^\Delta \neg b\, c^\Delta \neg f\, d^\Delta \neg k\, \neg l||\emptyset \Longrightarrow (Backchain\ False\ \lambda) \\
&a^\Delta \neg b\, c^\Delta \neg f\, d^\Delta \neg k\, \neg l\, \neg m||\emptyset \Longrightarrow (Unit\ Propagate\ LP) \\
&a^\Delta \neg b\, c^\Delta \neg f\, d^\Delta \neg k\, \neg l\, \neg m\, m||\emptyset
\end{aligned} \tag{26}$$

The construction in the proof of Lemma 8 applied to the nodes in this path gives following states of $\text{SML}_\Pi^\uparrow$:

$$\begin{aligned}
&\emptyset||\emptyset \\
&a^\Delta||\emptyset \\
&a^\Delta \neg b^{\neg b \vee \neg a}||\emptyset \\
&a^\Delta \neg b^{\neg b \vee \neg a}\, c^\Delta||\emptyset \\
&a^\Delta \neg b^{\neg b \vee \neg a}\, c^\Delta \neg f^{\neg f \vee \neg c}||\emptyset \\
&a^\Delta \neg b^{\neg b \vee \neg a}\, c^\Delta \neg f^{\neg f \vee \neg c}\, d^\Delta||\emptyset \\
&a^\Delta \neg b^{\neg b \vee \neg a}\, c^\Delta \neg f^{\neg f \vee \neg c}\, d^\Delta \neg k^{\neg k \vee \neg d}||\emptyset \\
&a^\Delta \neg b^{\neg b \vee \neg a}\, c^\Delta \neg f^{\neg f \vee \neg c}\, d^\Delta \neg k^{\neg k \vee \neg d}\, \neg l^{\neg l \vee b \vee k}||\emptyset \\
&a^\Delta \neg b^{\neg b \vee \neg a}\, c^\Delta \neg f^{\neg f \vee \neg c}\, d^\Delta \neg k^{\neg k \vee \neg d}\, \neg l^{\neg l \vee b \vee k}\, \neg m^{\neg m \vee l \vee b}||\emptyset \\
&a^\Delta \neg b^{\neg b \vee \neg a}\, c^\Delta \neg f^{\neg f \vee \neg c}\, d^\Delta \neg k^{\neg k \vee \neg d}\, \neg l^{\neg l \vee b \vee k}\, \neg m^{\neg m \vee l \vee b}\, m^{m \vee k \vee l}||\emptyset
\end{aligned} \tag{27}$$

It is clear that these nodes form a path in $\text{SML}_\Pi^\uparrow$ with every edge justified by the same transition rule as the corresponding edge in path (26) in $\text{SML}_\Pi$.

### 10.3 Proof of Proposition 11[↑]

In this section $\Pi$ is an arbitrary and fixed logic program.

For a record $M$, by $lcp(M)$ we denote its largest consistent prefix. We say that a clause $C$ is *conflicting* on a list $M$ of literals if $\Pi$ entails $C$, and $\overline{C} \subseteq lcp(M)$. For example, let $M$ be the first component of the last state in (27):

$$a^\Delta \neg b^{\neg b \vee \neg a} c^\Delta \neg f^{\neg f \vee \neg c} d^\Delta \neg k^{\neg k \vee \neg d} \neg l^{\neg l \vee b \vee k} \neg m^{\neg m \vee l \vee b} m^{m \vee k \vee l}. \tag{28}$$

Then, $lcp(M)$ is obtained by dropping the last element $m^{m \vee k \vee l}$ of $M$. It is clear that the reason $m \vee k \vee l$ for $m$ to be in $M$ is a conflicting clause on $M$.

*Lemma 10*

The literal that immediately follows $lcp(M)$ in an inconsistent record $M$, has the form $l^C$ where $C$ is a conflicting clause on $M$.

*Proof*

By the requirement (iii) of the definition of an extended record, the literal that immediately follows $lcp(M)$ may not be annotated by $\Delta$. Consequently, the literal has the form $l^C$. We now show that $C$ is a conflicting clause on $M$. Since $C$ is a reason for $l$ to be in $lcp(M) l^C$, it immediately follows that $\Pi$ entails $C$, $C$ can be written as $l \vee C'$, and $\overline{C'} \subseteq lcp(M)$. Since $l$ immediately follows the largest consistent prefix of $M$, $\overline{l} \in lcp(M)$. Consequently, $\overline{C} \subseteq lcp(M)$. We conclude that $C$ is indeed a conflicting clause on $M$. □

For any inconsistent record $l_1 \cdots l_n$ and any conflicting clause $C$ on this record, by $\beta_{l_1 \cdots l_n}(C)$ we denote the set of numbers $i$ such that $l_i \in \overline{C}$. (It is clear that every element from $\overline{C}$ equals to one of the literals in $l_1 \cdots l_n$.) The relation $I < J$ between subsets $I, J$ of $\{1 \cdots n\}$ is understood here as the lexicographical order between $I$ and $J$ sorted in descending order. For instance, $\{2\ 6\ 7\} < \{6\ 7\ 8\}$ because $\{7\ 6\ 2\} < \{8\ 7\ 6\}$ in lexicographical order.

Recall that the *resolution rule* can be applied to clauses $C \vee l$ and $C' \vee \neg l$ and produces the clause $C \vee C'$, called the *resolvent* of $C \vee l$ and $C' \vee \neg l$ on $l$.

*Lemma 11*

Let $M$ be a record and let $l^B$ be a nondecision literal from $lcp(M)$. If clause $D$ is the resolvent of $B$ and a clause $C$ conflicting on $M$ then

(i) $D$ is a clause conflicting on $M$,
(ii) $\beta_M(D) < \beta_M(C)$.

For instance, let $M$ be (28), let reason $\neg m \vee l \vee b$ for $\neg m$ in $lcp(M)$ be $B$, and let conflicting clause $m \vee k \vee l$ on $M$ be $C$. Then $D$, the result of resolving $B$ together with $C$, is clause $k \vee l \vee b$. Lemma 11 asserts that $k \vee l \vee b$ is a conflicting clause on $M$ and that $\beta_M(D) < \beta_M(C)$. Indeed, $\beta_M(D) = \{2\ 6\ 7\}$ and $\beta_M(C) = \{6\ 7\ 8\}$.

*Proof*
(i) Clause $D$ is a resolvent of $B$ and $C$ on some literal $l'$. Then, for some literal $l' \in B$, $\overline{l'} \in C$. The clause $C$ can be written as $\overline{l'} \vee C'$.

In order to demonstrate that $D$ is a conflicting clause we need to show that $\overline{D} \subseteq lcp(M)$ and $\Pi$ entails $D$.

Since $B$ is a reason for $l$ to be in $lcp(M)$, $\Pi$ entails $B$ and $B$ has the form $l \vee B'$ where $\overline{B'} \subseteq lcp(M)$. Since $C$ is a conflicting clause on $M$, $\overline{C} \subseteq lcp(M)$ and $\Pi$ entails $C$. From the fact that $lcp(M)$ is consistent, it follows that there is no literal in $B'$ such that its complement occurs in $C$. Consequently, $l' \notin B'$ so that $l'$ is $l$ and $D$ is $B' \vee C'$. We conclude that $\overline{D} \subseteq lcp(M)$. From the fact that $\Pi$ entails $B$, $\Pi$ entails $C$, and the construction of $D$, it follows that $\Pi$ entails $D$.

(ii) From the proof of (i) it follows that $D$ is a resolvent of $B$ and $C$ on $l$ where $B$ has the form $l \vee B'$. Since $B$ is a reason for $l$ to be in $lcp(M)$, every literal in $\overline{B'}$ precedes $l$ in $lcp(M)$. Since $D$ is derived by replacing $\overline{l}$ in $C$ with $B'$, $\beta_M(D) < \beta_M(B)$. □

Let record $M$ be $l_1 \cdots l_i \cdots l_n$, the *decision level* of a literal $l_i$ is the number of decision literals in $l_1 \cdots l_i$: we denote it by $dec_M(l_i)$. We will also use this notation to denote the decision level of a set of literals: For a set $P \subseteq M$ of literals, $dec_M(P)$ is the decision level of the literal in $P$ that occurs latest in $M$. For record $M$ and a decision level $j$ by $M^j$ we denote the prefix of $M$ that consists of the literals in $M$ that belong to decision level less than $j$ and by $M^{j]}$ we denote the prefix of $M$ that consists of the literals in $M$ that belong to decision level less than or equal to $j$. For instance, let $M$ be record (28) then $dec_M(\neg k) = 3$, $dec_M(\neg b\ c\ \neg k) = 3$, $M^3$ is $a^\Delta \neg b^{\neg b \vee \neg a} c^\Delta \neg f^{\neg f \vee \neg c}$, and $M^{3]}$ is $M$ itself.

*Lemma 12*
For an inconsistent record $M$ and a conflicting clause $l \vee C$ on $M$, if $dec_M(\overline{l}) > dec_M(\overline{c})$ for all $c \in C$ then $lcp(M)^{dec_M(\overline{C})]} l^{l \vee C}$ is a record.

*Proof*
We need to show that (i) $l \notin lcp(M)^{dec(\overline{C})]}$ and (ii) $l \vee C$ is a reason for $l$ to be in $lcp(M)^{dec(\overline{C})]}l$, i.e, $\Pi$ entails $l \vee C$ and $\overline{C} \subseteq lcp(M)^{dec(\overline{C})]}$.

Since $l \vee C$ is conflicting on $M$, $\overline{l \vee C} \subseteq lcp(M)$. From the consistency of $lcp(M)$ and the fact that $\overline{l} \in lcp(M)$, it follows that $l \notin lcp(M)$. Consequently, $l \notin lcp(M)^{dec(\overline{C})]}$.

Since $l \vee C$ is conflicting on $M$, $\Pi$ entails $l \vee C$ and $\overline{l \vee C} \subseteq lcp(M)$. Consequently, $\overline{C} \subseteq lcp(M)$. From the definition of $dec_M(\overline{C})$, it follows that $dec_M(\overline{C})$ is the decision level of the literal in $\overline{C}$ that occurs latest in $lcp(M)$. By the definition of a decision level, $\overline{C} \subseteq lcp(M)^{dec_M(\overline{C})]}$. □

*Proposition 11*$^\uparrow$
For a program $\Pi$, the transition rule *Backjump LP* is applicable to any backjump state in $\text{SML}_\Pi^\uparrow$.

*Proof*
Let $M||\Gamma$ be a backjump state in $\text{SML}_\Pi^\uparrow$. Let $R$ be the list of reasons that are assigned to the nondecision literals in $lcp(M)$.

Consider the process of building a sequence $C_1, C_2, \ldots$ of clauses so that

- $C_1$ is the reason of the member of $M$ that immediately follows $lcp(M)$, and
- $C_j$ $(j > 1)$ is a resolvent of $C_{j-1}$ and some clause in $R$

while derivation of new clauses is possible. From Lemma 11 (i) and the choice of $C_1$ and $R$, it follows that any clause in $C_1, C_2 \ldots$ is conflicting. By Lemma 11 (ii) we conclude that $\beta_M(C_j) < \beta_M(C_{j-1})$ $(j > 1)$. It is clear that this process will terminate after deriving some clause $C_m$, since the number of conflicting clauses on $M$ is finite. It is clear that clause $C_m$ cannot be resolved against any clause in $R$.

Case 1. $C_m$ is the empty clause. Since $M||\Gamma$ is a backjump state, $M$ contains a decision literal $l^\Delta$. By part (iii) of the definition of a record, $l$ belongs to $lcp(M)$. Consequently, $M$ can be represented in the form $lcp(M)^{dec_M(l)} l^\Delta Q$.

By the choice of $C_1$, $C_1$ is a reason and must consist of at least one literal. Consequently, $m > 1$. Clause $C_m$ is derived from clauses $C_{m-1}$ and some clause in $R$. Since $C_m$ is empty, $C_{m-1}$ is a unit clause $l'$. We will show that

$$lcp(M)^{dec_M(l)} l^\Delta Q||\Gamma \Longrightarrow lcp(M)^{dec_M(l)} l'^{l'}||\Gamma$$

is an application of *Backjump LP*. It is sufficient to demonstrate that $lcp(M)^{dec_M(l)} l'^{l'}$ is a record. Since $lcp(M)^{dec_M(l)} l^\Delta Q$ is a record, we only need to show that $l' \notin lcp(M)^{dec_M(l)}$ and clause $l'$ is a reason for $l'$ to be in $lcp(M)^{dec_M(l)} l'$. Recall that $C_{m-1}$, i.e., $l'$, is a conflicting clause. Consequently, $\Pi$ entails $l'$ and $\overline{l'} \in lcp(M)$. Since $lcp(M)$ is consistent, $l' \notin lcp(M)$ so that $l' \notin lcp(M)^{dec_M(l)}$. On the other hand, from the fact that $\Pi$ entails $l'$ it immediately follows that clause $l'$ is a reason for $l'$ to be in $lcp(M)^{dec_M(l)} l'$.

Case 2. $C_m$ is not empty. Since $C_m$ is a conflicting clause on $M$, the complement of any literal in $C_m$ belongs to $lcp(M)$. Furthermore, every such complement is a decision literal in $lcp(M)$. Indeed, if this complement is $\overline{l}^{\overline{l} \vee B} \in lcp(M)$ then $\overline{l} \vee B$ is one of the clauses $B_i$, and it can be resolved against $C_m$.

By the definition of a decision level, there is at most one decision literal that belongs to any decision level. It follows that $C_m$ can be written as $l \vee C'_m$ so that $dec_M(\overline{l}) > dec_M(\overline{c})$ for any $c \in C'_m$. Consequently, $M$ can be written as $lcp(M)^{dec_M(\overline{l})} \overline{l}^\Delta Q$. Note that

$$lcp(M)^{dec_M(\overline{l})} \overline{l}^\Delta Q||\Gamma \Longrightarrow lcp(M)^{dec_M(\overline{C'_m})]} l^{C_m}||\Gamma$$

is an application of *Backjump LP*. Indeed, by Lemma 12 $lcp(M)^{dec_M(\overline{C'_m})]} l^{C_m}$ is a record. □

Algorithm 1 presents procedure *BackjumpClause* that computes a backjump clause for any backjump state in the graph $\text{SML}_\Pi^\uparrow$. The algorithm follows from the construction of the proof of Proposition 11$^\uparrow$. It is based on the iterative application of the resolution rule on reasons of the smallest inconsistent prefix of a state. The

```
BackjumpClause (M||Γ);
Arguments    : M||Γ is a backjump state in SML↑_Π
Return Value : C is a backjump clause
begin
    C ← the reason of the member of M that immediately follows lcp(M);
    N ← the list of the nondecision literals in lcp(M);
    R ← the list of the reasons that are assigned to the literals in N;
    while C̄ ∩ N ≠ ∅ do
        l ← a literal in C̄ ∩ N;
        B ← the clause in R that contains l;
        C' ← the resolvent of C and B on l;
        if C' = ∅ then
          ⌊ return C
        C ← C'
    return C ;
```

**Algorithm 1:** A procedure for generating a backjump clause.

proof of Proposition 11$^\uparrow$ allows to conclude the termination of *BackjumpClause* and asserts that a clause returned by the procedure is a backjump clause on a backjump state.

For instance, let $\Pi$ be (25). Consider an execution of *BackjumpClause* on $\Pi$ and backjump state (28). The table below gives the values of $lcp(M)$, $C$, $N$, and $R$ during the execution of the *BackjumpClause* algorithm. By $C_i$ we denote a value of $C$ before the $i$-th iteration of the **while** loop.

$$
\begin{array}{l|l}
lcp(M) & a^\Delta \neg b^{\neg b \vee \neg a} c^\Delta \neg f^{\neg f \vee \neg c} d^\Delta \neg k^{\neg k \vee \neg d} \neg l^{\neg l \vee b \vee k} \neg m^{\neg m \vee l \vee b} \\
C_1 & m \vee k \vee l \\
N & \neg b^{\neg b \vee \neg a} \neg f^{\neg f \vee \neg c} \neg k^{\neg k \vee \neg d} \neg l^{\neg l \vee b \vee k} \neg m^{\neg m \vee l \vee b} \\
R & \neg b \vee \neg a,\ \neg f \vee \neg c,\ \neg k \vee \neg d,\ \neg l \vee b \vee k,\ \neg m \vee l \vee b \\
\hline
C_2 & k \vee l \vee b \text{ is the resolvent of } C_1 \text{ and } \neg m \vee l \vee b \\
C_3 & k \vee b \text{ is the resolvent of } C_2 \text{ and } \neg l \vee b \vee k \\
C_4 & \neg d \vee b \text{ is the resolvent of } C_3 \text{ and } \neg k \vee \neg d \\
C_5 & \neg d \vee \neg a \text{ is the resolvent of } C_4 \text{ and } \neg b \vee \neg a
\end{array}
\qquad (29)
$$

The algorithm will terminate with the clause $\neg d \vee \neg a$. Proof of Proposition 11$^\uparrow$ asserts that (i) this clause is a backjump clause such that $d$ and $a$ are decision literals in $M$ and (ii) the transition

$$
\begin{array}{l}
a^\Delta \neg b^{\neg b \vee \neg a} c^\Delta \neg f^{\neg f \vee \neg c} d^\Delta \neg k^{\neg k \vee \neg d} \neg l^{\neg l \vee b \vee k} \neg m^{\neg m \vee l \vee b} m^{m \vee k \vee l} ||\emptyset \Longrightarrow \\
a^\Delta \neg b^{\neg b \vee \neg a} \neg d^{\neg d \vee \neg a} ||\emptyset
\end{array}
\qquad (30)
$$

in SML$^\uparrow_\Pi$ is an application of *Backjump LP*. Indeed, by Lemma 12 $lcp(M)^{dec_M(\overline{\neg a})]} \neg d^{\neg d \vee \neg a}$, in other words $a^\Delta \neg b^{\neg b \vee \neg a} \neg d^{\neg d \vee \neg a}$, is a record.

Note that a backjump clause may be derived in other ways than captured by *BackjumpClause* algorithm: the transition rule *Backjump LP* is applicable with an arbitrary backjump clause. Usually, DPLL-like procedures implement conflict-driven backjumping and learning where a particular learning schema such as, for instance, *Decision* or *FirstUIP* (Mitchell 2005) is applied for computing a special kind of a backjump clause. It turns out that the *BackjumpClause* algorithm captures the *Decision* learning schema for ASP. Typically, SAT solvers impose an order for resolving the literals during the process of *Decision* backjump clause derivation. We can impose similar order by replacing the line

$$l \leftarrow \text{a literal in } \overline{C} \cap N$$

in the algorithm *BackjumpClause* with

$$l \leftarrow \text{a literal in } \overline{C} \cap N \text{ that occurs latest in } lcp(M).$$

In fact, the sample application of *BackjumpClause* algorithm described in (29) follows this ordering.

This section introduced *BackjumpClause* algorithm that derives a *Decision* backjump clause for an arbitrary backjump state. In the next section we will introduce an algorithm that will compute an ASP counterpart of *FirstUIP* backjump clause.

## 11 FirstUIP Conflict-Driven Backjumping and Learning

Conflict-driven backjumping and learning proved to be a highly successful technique in modern SAT solving. Furthermore, in (Zhang et al. 2001) the authors investigated the performance of various learning schemes and established experimentally that *FirstUIP* clause is the most useful single clause to learn. Success of conflict-driven learning led to the implementation of its ASP counterpart in systems SMODELS$_{cc}$, CLASP, and SUP. There are two common methods for describing a backjump clause construction in the SAT literature. The first one employs the implication graph (Marques-Silva and Sakallah 1996) and the second one employs resolution (Mitchell 2005). Ward and Schlipf (Ward and Schlipf 2004) extended the definition of an implication graph to the SMODELS algorithm and implemented *FirstUIP* learning schema in answer set solver SMODELS$_{cc}$. In the previous section we used SML$_\Pi^\uparrow$ formalism and resolution to describe the *BackjumpClause* algorithm for computing an ASP counterpart of a *Decision* backjump clause. In (Gebser et al. 2007) the authors used the concepts from constraint processing to implement *FirstUIP* learning schema in answer set solver CLASP.

This section presents the *BackjumpClauseFirstUIP* algorithm for computing an ASP counterpart of a *FirstUIP* backjump clause by means of SML$_\Pi^\uparrow$ formalism and resolution. The *BackjumpClauseFirstUIP* algorithm is employed by the system SUP in its implementation of conflict-driven backjumping and learning.

The Algorithm 2 presents procedure *BackjumpClauseFirstUIP* that computes a *FirstUIP* backjump clause for any backjump state in the graph SML$_\Pi^\uparrow$.

We now state the correctness of the algorithm *BackjumpClauseFirstUIP*. We start by showing its termination. By $C_1$ we will denote the initial value assigned to clause $C$. From Lemma 11 (i) and the choice of $C_1$ we conclude that at any point of computation clause $C$ is conflicting on $M$. By Lemma 11 (ii), the value of $\beta_M(C)$ decreases with each new assignment of clause $C$ in the **while** loop. It follows that the **while** loop will terminate since the number of conflicting clauses $C$ on $M$ such

```
BackjumpClauseFirstUIP (M||Γ);
```
**Arguments** : $M||\Gamma$ is a backjump state in $\text{SML}_\Pi^\uparrow$
**Return Value** : $C$ is a backjump clause
**begin**

    $C \leftarrow$ the reason of the member of $M$ that immediately follows $lcp(M)$;
    $l \leftarrow$ the literal in $\overline{C}$ that occurs latest in $lcp(M)$;
    $P \leftarrow$ the sublist of $lcp(M)$ that consists of the literals that belong to the decision level $dec(l)$;
    $R \leftarrow$ the list of the reasons that are assigned to the literals in $P$;
    **while** $|\overline{C} \cap P| > 1$ **do**
        $l \leftarrow$ the literal in $\overline{C}$ that occurs latest in $P$;
        $B \leftarrow$ the clause in $R$ that contains $l$;
        $C \leftarrow$ the resolvent of $C$ and $B$ on $l$;
    **return** $C$ ;

**Algorithm 2:** A procedure for generating a FirstUIP backjump clause.

that $|\overline{C} \cap P| > 1$ is finite. By $C_m$ we will denote the clause $C$ with which the **while** loop terminates. In other words *BackjumpClauseFirstUIP* returns $C_m$. We now show that $C_m$ is indeed a backjump clause. We already concluded that $C_m$ is a conflicting clause on $M$. Furthermore, from the termination condition of the **while** loop $|\overline{C_m} \cap P| \leq 1$. From the choice of $C_1$ and $P$ it follows that $|\overline{C_m} \cap P| = 1$. Consequently, $C_m$ can be written as $l \vee C'_m$ where $\bar{l}$ is in singleton $\overline{C_m} \cap P$. By Lemma 11 (ii), $\beta(C_m) \leq \beta(C_1)$. From the definition of $\beta$ and the choice of $P$ it follows that $dec_M(\bar{l}) > dec_M(\bar{c})$ for all $c \in C'_m$. By Lemma 12, $lcp(M)^{dec_M(\overline{C'_m})]} l^{C_m}$ is a record. In other words, transition

$$M||\Gamma \Longrightarrow lcp(M)^{dec_M(\overline{C'_m})]} l^{C_m} ||\Gamma$$

is an application of *Backjump LP*. Consequently, $C_m$ is a backjump clause.

For instance, let $\Pi$ be (25). Consider an execution of *BackjumpClauseFirstUIP* on $\Pi$ and a backjump state (28). The table below gives the values of $lcp(M)$, $C$, $P$, and $R$ during the execution of *BackjumpClauseFirstUIP*. By $C_i$ we denote a value of $C$ before the $i$-th iteration of the **while** loop.

| | |
|---|---|
| $lcp(M)$ | $a^\Delta \neg b^{\neg b \vee \neg a} c^\Delta \neg f^{\neg f \vee \neg c} d^\Delta \neg k^{\neg k \vee \neg d} \neg l^{\neg l \vee b \vee k} \neg m^{\neg m \vee l \vee b}$ |
| $C_1$ | $m \vee k \vee l$ |
| $P$ | $d^\Delta \neg k^{\neg k \vee \neg d} \neg l^{\neg l \vee b \vee k} \neg m^{\neg m \vee l \vee b}$ |
| $R$ | $\neg k \vee \neg d,\ \neg l \vee b \vee k,\ \neg m \vee l \vee b$ |
| $C_2$ | $k \vee l \vee b$ is the resolvent of $C_1$ and $\neg m \vee l \vee b$ |
| $C_3$ | $k \vee b$ is the resolvent of $C_2$ and $\neg l \vee b \vee k$. |

The *BackjumpClauseFirstUIP* algorithm will terminate with the clause $k \vee b$. The proof of the correctness of *BackjumpClauseFirstUIP* asserts that $k \vee b$ is a backjump clause and the transition

$$a^\Delta \neg b^{\neg b \vee \neg a} c^\Delta \neg f^{\neg f \vee \neg c} d^\Delta \neg k^{\neg k \vee \neg d} \neg l^{\neg l \vee b \vee k} \neg m^{\neg m \vee l \vee b} m^{m \vee k \vee l} \Longrightarrow$$
$$a^\Delta \neg b^{\neg b \vee \neg a} k^{k \vee b} || \emptyset \qquad (31)$$

in $\text{SML}^{\uparrow}_\Pi$ is an application of *Backjump LP*.

## 12 Extended Graph: Generate and Test

In this section we introduce an extended graph $\text{GTL}^{\uparrow}_{F,G}$ for the *generate and test* abstract framework $\text{GTL}_{F,G}$ similar as in Section 9 we introduced $\text{SML}^{\uparrow}_\Pi$ for $\text{SML}_\Pi$.

For a formula $H$, we say that a clause $l \vee C$ is a *reason* for $l$ to be in a list $P\, l\, Q$ of literals w.r.t. $H$ if $H \models l \vee C$ and $\overline{C} \subseteq P$.

An *(extended) record* $M$ relative to a formula $H$ is a list of literals over the set of atoms occurring in $H$ where

(i) each literal $l$ in $M$ is annotated either by $\Delta$ or by a reason for $l$ to be in $M$ w.r.t. $H$,
(ii) $M$ contains no repetitions,
(iii) for any inconsistent prefix of $M$ its last literal is annotated by a reason.

An *(extended) state* relative to a CNF formula $F$, and a formula $G$ formed from atoms occurring in $F$ is either a distinguished state *FailState* or a pair of the form $M||\Gamma$, where $M$ is an extended record relative to $F \wedge G$, and $\Gamma$ is the same as in the definition of an augmented state (i.e., $\Gamma$ is a (multi-)set of clauses formed from atoms occurring in $F$ that are entailed by $F \wedge G$.) For any extended state $S$ relative to $F$ and $G$, the result of removing annotations from all nondecision literals of $S$ is a state of $\text{GTL}_{F,G}$: we will denote this state by $S^{\downarrow}$.

For a CNF formula $F$ and a formula $G$ formed from atoms occurring in $F$, we will define a graph $\text{GTL}^{\uparrow}_{F,G}$. The set of the nodes of $\text{GTL}^{\uparrow}_{F,G}$ consists of the extended states relative to $F$ and $G$. The transition rules of $\text{GTL}_{F,G}$ are extended to $\text{GTL}^{\uparrow}_{F,G}$ as follows: $S_1 \Longrightarrow S_2$ is an edge in $\text{GTL}^{\uparrow}_{F,G}$ justified by a transition rule $T$ if and only if $S_1^{\downarrow} \Longrightarrow S_2^{\downarrow}$ is an edge in $\text{GTL}_{F,G}$ justified by $T$.

The lemma below formally states the relationship between nodes of the graphs $\text{GTL}_{F,G}$ and $\text{GTL}^{\uparrow}_{F,G}$:

*Lemma 13*
For any CNF formula $F$ and a formula $G$ formed from atoms occurring in $F$, if $S'$ is a state reachable from $\emptyset||\emptyset$ in the graph $\text{GTL}_{F,G}$ then there is a state $S$ in the graph $\text{GTL}^{\uparrow}_{F,G}$ such that $S^{\downarrow} = S'$.

The definitions of Basic transition rules and semi-terminal states in $\text{GTL}^{\uparrow}_{F,G}$ are similar to their definitions for $\text{GTL}_{F,G}$.

*Proposition 9$^{\uparrow}$*
For any CNF formula $F$ and a formula $G$ formed from atoms occurring in $F$,

(a) every path in $\text{GTL}^{\uparrow}_{F,G}$ contains only finitely many edges labeled by Basic transition rules,
(b) for any semi-terminal state $M||\Gamma$ of $\text{GTL}^{\uparrow}_{F,G}$, $M$ is a model of $F \wedge G$,
(c) $\text{GTL}^{\uparrow}_{F,G}$ contains an edge leading to *FailState* if and only if $F \wedge G$ is unsatisfiable.

We say that a state in the graph $\text{GTL}_{F,G}^\uparrow$ is a *backjump state* if its record is inconsistent and contains a decision literal. As in case of the graph $\text{GTL}_{F,G}$, any backjump state in $\text{GTL}_{F,G}^\uparrow$ is not semi-terminal:

*Proposition 10$^\uparrow$*

For any CNF formula $F$ and a formula $G$ formed from atoms occurring in $F$, the transition rule *Backjump GT* is applicable in any backjump state in $\text{GTL}_{F,G}^\uparrow$.

Proposition 9 (b), (c) and Proposition 10 easily follow from Lemma 13 and Proposition $9^\uparrow$ (b), (c) and Proposition $10^\uparrow$ respectively. Proof of Proposition 9 (a) is similar to the proof of Proposition $9^\uparrow$ (a).

## 13 Proofs of Proposition $9^\uparrow$, Lemma 13, Proposition $10^\uparrow$

### 13.1 Proof of Proposition $9^\uparrow$

*Lemma 14*

For any CNF formula $F$, a formula $G$ formed from atoms occurring in $F$, an extended record $M$ relative to $F \wedge G$, and any model $X$ of $F \wedge G$, if $X$ satisfies all decision literals in $M$ then $X \models M$.

*Proof*

By induction on the length of $M$. The property trivially holds for $\emptyset$. We assume that the property holds for any state with $n$ elements. Consider any state $M$ with $n+1$ elements. Let $X$ be a model of $F \wedge G$ such that $X$ satisfies all decision literals in $M$.

Case 1. $M$ has the form $P\,l^\Delta$. By the inductive hypothesis, $X \models P$. Since $X$ satisfies all decision literals in $M$, $X \models l^\Delta$.

Case 2. $M$ has the form $P\,l^{l \vee C}$. By the inductive hypothesis, $X \models P$. By the definition of a reason (i) $F \wedge G$ entails $l \vee C$ and (ii) $\overline{C} \subseteq P$. From (ii) it follows that $P \models \neg C$. Consequently, $X \models \neg C$. From (i) it follows that $X \models l \vee C$. We conclude that $X \models l$.  $\square$

The proof of Proposition $9^\uparrow$ assumes the correctness of Proposition $10^\uparrow$ that we demonstrate in Section 13.3.

ıProof of Proposition $9^\uparrow$

Parts (a) and (c) are proved as in the proof of Proposition $8^\uparrow$, using Lemma 14.
(b) Let $M||\Gamma$ be a semi-terminal state so that none of the Basic rules are applicable. From the fact that *Decide* is not applicable, we conclude that $M$ assigns all literals.

Furthermore, $M$ is consistent. Indeed, assume that $M$ is inconsistent. Then, since *Fail* is not applicable, $M$ contains a decision literal. Consequently, $M||\Gamma$ is a backjump state. By Proposition $10^\uparrow$, the transition rule *Backjump GT* is applicable in $M||\Gamma$. This contradicts our assumption that $M||\Gamma$ is semi-terminal.

Also, $M$ is a model of $F$: since *Unit Propagate* $\lambda$ is not applicable, it follows that for every clause $C \vee l \in F \cup \Gamma$ if $\overline{C} \subseteq M$ then $l \in M$. Consequently, $M \models C \vee l$. Furthermore, $M$ is a model of $G$: since *Test* is not applicable, then $G \not\models \overline{M}$. We conclude that $M \models G$. Consequently, $M$ is a model of $F \wedge G$.  $\square$

### 13.2 Proof of Lemma 13

For a state $S$ in the graph $\text{GTL}^{\uparrow}_{F,G}$, we say that $S^{\downarrow}$ in $\text{GTL}_{F,G}$ is the *image* of $S$.

*Lemma 13*
For any CNF formula $F$ and a formula $G$ formed from atoms occurring in $F$, if $S'$ is a state reachable from $\emptyset||\emptyset$ in the graph $\text{GTL}_{F,G}$ then there is a state $S$ in the graph $\text{GTL}^{\uparrow}_{F,G}$ such that $S^{\downarrow} = S'$.

*Proof*
Since the property trivially holds for the initial state $\emptyset||\emptyset$, we only need to prove that all transition rules of $\text{GTL}_{F,G}$ preserve it.

Consider an edge $M||\Gamma \implies M'||\Gamma'$ in the graph $\text{GTL}_{F,G}$ such that there is a state $M_1||\Gamma$ in the graph $\text{GTL}^{\uparrow}_{F,G}$ satisfying the condition $(M_1||\Gamma)^{\downarrow} = M||\Gamma$. We need to show that there is a state in the graph $\text{GTL}^{\uparrow}_{F,G}$ such that $M'||\Gamma'$ is its image in $\text{GTL}_{F,G}$. Consider several cases that correspond to a transition rule leading from $M||\Gamma$ to $M'||\Gamma'$:

*Unit Propagate* $\lambda$:

$$M||\Gamma \implies M\,l||\Gamma \quad \text{if} \quad \begin{cases} C \vee l \in F \cup \Gamma \text{ and} \\ \overline{C} \subseteq M. \end{cases}$$

$M'||\Gamma'$ is $M\,l||\Gamma$. It is sufficient to prove that $M_1\,l^{C \vee l}||\Gamma$ is a state of $\text{GTL}^{\uparrow}_{F,G}$. It is enough to show that a clause $C \vee l$ is a reason for $l$ to be in $M\,l$ w.r.t. $F \wedge G$, i.e, $F \wedge G \models C \vee l$ and $\overline{C} \subseteq M$. By applicability conditions of *Unit Propagate* $\lambda$, $\overline{C} \subseteq M$. By the definition of a state $F \wedge G$ entails $\Gamma$. Since $C \vee l \in F \cap \Gamma$, $F \wedge G \models C \vee l$.

*Test*:

$$M||\Gamma \implies M\overline{l}||\Gamma \quad \text{if} \quad \begin{cases} M \text{ is consistent,} \\ G \models \overline{M}, \\ l \in M. \end{cases}$$

$M'||\Gamma'$ is $M\overline{l}||\Gamma$. It is sufficient to prove that $M_1\,\overline{l}^{\overline{M}}||\Gamma$ is a state of $\text{GTL}^{\uparrow}_{F,G}$. $\overline{M}$ has the form $\overline{l} \vee C$. It is enough to show that a clause $\overline{l} \vee C$ is a reason for $\overline{l}$ to be in $M\overline{l}$ w.r.t. $F \wedge G$. It is trivial that $\overline{C} \subseteq M$. By applicability condition of the rule, $G \models \overline{l} \vee C$.

*Backjump GT*, *Decide*, *Fail*, and *Learn GT*: obvious. $\square$

### 13.3 Proof of Proposition 10$^{\uparrow}$

For a state $M\,l^C||\Gamma$, we say that a reason $C$ is a *backjump clause* if there is a transition *Backjump GT* leading to $M\,l^C||\Gamma$ in $\text{GTL}_{F,G}$.

In this section $F$ is an arbitrary and fixed CNF formula and $G$ is an arbitrary and fixed formula formed from atoms occurring in $F$.

For a record $M$, by $lcp(M)$ we denote its largest consistent prefix. We say that a clause $C$ is *conflicting* on a list $M$ of literals if $F \wedge G$ entails $C$, and $\overline{C} \subseteq lcp(M)$.

Lemmas 10, 11, 12 hold for the case of extended record relative to a formula. The proofs of the lemmas have to be modified only by replacing $\Pi$ with $F \wedge G$.

Proposition 10$^{\uparrow}$ is proved as Proposition 11$^{\uparrow}$.

Algorithms *BackjumpClause* and *BackjumpClauseFirstUIP* are applicable to the backjump states of the graph $\text{GTL}^{\uparrow}_{F,G}$.

## 14 Experiments with Sup

Here we present experimental analysis that compares performance of the system SUP versus CMODELS, CLASP, SMODELS, and SMODELS$_{cc}$. We start by describing the implementation details of SUP.

The implementation of SUP utilizes

- the interface of SAT-solver MINISAT (v1.12b) that supports non-clausal constraints described in (Een and Sörensson 2003) in order to introduce additional inference possibilities, but unit propagation. In particular, SUP implements *Backchain True* and *All Rules Cancelled* by means of non-clausal constraints and it uses the unit propagate of MINISAT to capture *Unit Propagate LP* and *Backchain False*.
- parts of CMODELS code that eliminate weight and choice rules; perform model verification; and compute loop formulas. In particular, SUP uses the latter two parts of CMODELS code to capture *Unfounded*.

In the experiments we used the following versions of the systems: SUP v. 0.1, SUP v. 0.2, CMODELS v. 3.77 using MINISAT v. 1.12b, CLASP v. 1.0.5, SMODELS v. 2.32, SMODELS$_{cc}$ v. 1.08 (implemented on top of SMODELS v. 2.26). System SUP (v. 0.1 and v. 0.2) extends the implementation of MINISAT v. 1.12b. Therefore, we compare SUP performance against CMODELS that uses MINISAT 1.12b for its inference. System SUP v. 0.1 stands for a version of SUP that implements *Unit Propagate LP*, *All Rules Cancelled*, and *Backchain False* $\lambda$ propagation rules, and does not implement *Backchain True*. System SUP v 0.2, on the other hand, also implements *Backchain True*.

All considered solvers use preprocessor LPARSE (see Footnote 5) to ground the problems so that the systems are run on identical ground instances. Grounding time is not accounted for in solving time. All times are reported in seconds. Symbol *tout* stands for the fact that a system did not terminate with a solution after 10 minutes. *Sup 0.1, Sup, Cm, Cl, Sm$_{cc}$,* and *Sm* stand for SUP v. 0.1, SUP v. 0.2, CMODELS, CLASP, SMODELS$_{cc}$, and SMODELS respectively. The symbol $-t$ abbreviates the flag $-temp$ that allows SUP to forget learnt clauses due to loop formulas (by default SUP adds these clauses into permanent clause database). The symbol $-a$ abbreviates the flag $-atomreason$ that forces CMODELS, like SUP, to add only a clause implied by some loop formula and unsatisfied by a current model rather than the complete loop formula unsatisfied by the model. By default, CMODELS adds a complete loop formula unsatisfied by the model. All experiments were run on Intel(R) Pentium(R) D CPU 3.00GHz, 2 cpu cores, cache size 1024 KB, running Linux.

Table 1 presents the experiments run on tight programs. Recall that for tight programs (i) the transition rule *Unfounded* of SUP is never used for inference and (ii) the transition rule *Test* of CMODELS is never used for inference.

| Instance | Sup 0.1 | Sup | Cm | Cl | $Sm_{cc}$ | Sm |
|---:|---|---|---|---|---|---|
| Towers of Hanoi: http://asparagus.cs.uni-potsdam.de ||||||  |
| towers-hanoi.35_6 | 43.68 | 92.96 | 74.03 | 23.86 | 115.83 | 62.19 |
| towers-hanoi.36_6 | 183.05 | 70.71 | 117.56 | 38.98 | 112.44 | 86.01 |
| towers-hanoi.37_6 | 35.71 | 30.75 | 290.65 | 34.25 | 84.75 | 120.53 |
| towers-hanoi.38_6 | 243.41 | 233.82 | 37.03 | 70.94 | 99.87 | 168.88 |
| towers-hanoi.39_6 | 96.59 | tout | tout | 123.40 | 384.24 | 237.90 |
| towers-hanoi.40_6 | tout | 113.15 | 30.21 | 114.39 | 124.73 | 329.08 |
| towers-hanoi.41_6 | 123.00 | 69.51 | 103.61 | 169.43 | 168.74 | 466.15 |
| towers-hanoi.42_6 | tout | 389.88 | 91.28 | 182.10 | tout | tout |
| towers-hanoi.43_6 | tout | 42.40 | 353.74 | 228.37 | 204.89 | tout |
| towers-hanoi.44_6 | 501.80 | 438.78 | 498.89 | tout | tout | tout |
| Pigeon Holes with 10 holes: pgh#pigeons ||||||  |
| pgh7 | 0.18 | 0.22 | 0.14 | 0.10 | 20.98 | 1.67 |
| pgh8 | 0.75 | 1.10 | 0.68 | 1.00 | tout | 8.87 |
| pgh9 | 7.60 | 9.30 | 4.03 | 5.73 | tout | 47.31 |
| Queens Normal Encoding: q.lp.#queens ||||||  |
| q.lp.18 | 0.14 | 0.14 | 0.14 | 0.08 | 155.62 | 7.84 |
| q.lp.22 | 0.28 | 0.27 | 0.30 | 0.14 | tout | tout |
| q.lp.24 | 0.35 | 0.35 | 0.38 | 0.20 | tout | tout |
| q.lp.30 | 0.70 | 0.71 | 0.75 | 0.44 | tout | tout |
| Queens Cardinality Constraint Encoding: q.lp2.#queens ||||||  |
| q.lp2.18 | 0.06 | 0.05 | 0.06 | 0.04 | 147.88 | 2.24 |
| q.lp2.22 | 0.11 | 0.10 | 0.12 | 0.08 | tout | 191.76 |
| q.lp2.24 | 0.15 | 0.14 | 0.17 | 0.11 | tout | 267.27 |
| q.lp2.30 | 0.30 | 0.28 | 0.32 | 0.21 | tout | tout |
| TOAST: http://asparagus.cs.uni-potsdam.de ||||||  |
| sequence3-ss3-Plain | 138.08 | 49.69 | 78.45 | 10.51 | 444.41 | tout |
| sequence4-ss2-Plain | 14.86 | 14.68 | 24.50 | 9.56 | 99.85 | 71.42 |
| sequence4-ss3-Plain | 408.99 | 468.25 | tout | 294.27 | tout | tout |
| sequence3-ss2 | 8.11 | 9.00 | 8.34 | 5.75 | 79.23 | 46.07 |
| sequence3-ss3 | 137.37 | 49.84 | 78.45 | 10.39 | 444.16 | tout |
| sequence4-ss2 | 16.13 | 16.88 | 13.38 | 8.96 | 102.98 | 74.85 |
| sequence4-ss3 | 103.33 | 207.21 | 16.74 | 233.09 | tout | tout |
| Vertex Cover vcx.# minimum size vertex cover: http://www.cs.engr.uky.edu/ai/benchmarks.html ||||||  |
| vc1.53 | 8.30 | 3.23 | 19.81 | tout | tout | 218.59 |
| vc2.50 | 11.55 | 6.76 | 13.32 | 94.43 | tout | 11.73 |
| vc3.55 | 0.65 | 4.09 | 64.44 | 512.56 | tout | 10.59 |
| vc4.54 | 6.49 | 10.19 | 23.05 | tout | tout | tout |

Table 1: Experiments: Tight problems.

Table 2 presents the experiments run on nontight programs:

| Instance | Sup | Sup -t | Cm | Cm -a | Cl | Sm$_{cc}$ | Sm |
|---|---|---|---|---|---|---|---|
| Deterministic Automaton: http://www.fmi.uni-stuttgart.de/szs/ | | | | | | | |
| research/projects/synthesis/benchmarks030923.html | | | | | | | |
| mutex3Morin | 15.30 | 15.25 | 15.68 | 15.68 | 15.45 | 306.30 | 153.60 |
| mutex4IDFD | 0.80 | 0.74 | 0.66 | 0.64 | 0.80 | 35.53 | 13.46 |
| phi3Morin | 0.96 | 0.95 | 0.62 | 0.72 | 0.37 | 2.50 | 2.80 |
| phi4IDFD | 12.11 | 14.85 | 0.39 | 2.98 | 0.02 | 0.31 | 0.16 |
| phi4Morin | tout | 448.82 | 105.54 | tout | 95.50 | tout | tout |
| phi5IDFD | tout | tout | 67.14 | tout | 138.56 | tout | tout |
| Bounded Model Checking: | | | | | | | |
| http://www.tcs.hut.fi/~kepa/experiments/boundsmodels/ | | | | | | | |
| dp10.i.O2.b12 | 30.88 | 3.51 | 10.14 | 22.29 | 0.20 | tout | 63.11 |
| dp10.s.O2.b9 | 0.54 | 0.42 | 0.78 | 0.68 | 0.14 | 26.18 | 13.04 |
| dp12.i.O2.b14 | 254.79 | 88.18 | 188.89 | 272.55 | 7.17 | tout | tout |
| dp12.s.O2.b10 | 5.89 | 2.01 | 0.76 | 2.05 | 0.72 | tout | 337.28 |
| dp6.i.O2.b8 | 0.20 | 0.20 | 0.18 | 0.16 | 0.02 | 0.93 | 0.32 |
| dp8.i.O2.b10 | 0.98 | 0.77 | 1.53 | 2.60 | 0.03 | 12.97 | 4.12 |
| dp8.s.O2.b8 | 0.12 | 0.20 | 0.15 | 0.10 | 0.02 | 2.58 | 1.18 |
| Hamiltonian Cycle: http://www.cs.engr.uky.edu/ai/benchmarks.html | | | | | | | |
| hc_1S | tout | 2.82 | tout | tout | tout | tout | tout |
| hc_2S | 0.29 | 5.37 | 13.50 | 8.60 | 0.38 | 153.44 | tout |
| hc_3S | 1.28 | 8.15 | 5.94 | 3.10 | tout | tout | tout |
| hc_4S | 7.08 | 2.81 | tout | 0.94 | 2.18 | 14.92 | tout |

Table 2: Experiments: Nontight problems.

Overall the results demonstrated by SUP place the system in the class of highly efficient answer set solvers.

## 15 Related Work

Simons (2000) and Ward (2004) described the SMODELS and SMODELS$_{cc}$ algorithms, respectively, by means of pseudocode and demonstrated their correctness. In this paper we designed an abstract framework that was used as an alternative method for describing these algorithms and demonstrating their correctness.

Gebser and Schaub (2006) provided a deductive system for describing inferences involved in computing answer sets by tableaux methods. The abstract framework presented here can be viewed as a deductive system also, but of a very different kind. First, it accounts for phenomena such as backjumping and learning (and also forgetting and restart) whereas the Gebser-Schaub system does not. Second, we describe backtracking by an inference rule, and the Gebser-Schaub system does not. Accordingly, the derivations considered in this paper describe search process, and derivations in the Gebser-Schaub system do not. Also, the abstract framework discussed here does not have any inference rule similar to Cut; this is why its derivations are paths, rather than trees.

## 16 Conclusions

In this paper we showed how to model advanced algorithms for computing answer sets of a program by means of simple mathematical objects, graphs. We extended the abstract frameworks proposed in (Lierler 2008) for describing native and SAT-based ASP algorithms to capture such sophisticated features as backjumping and learning. We characterized the algorithms of systems SMODELS$_{cc}$, SUP, and CMODELS that implement these features. We note that the work on this abstract framework suggested the implementation of answer set solver SUP and the experimental analysis presented here demonstrates that SUP is a competitive representative in the family of answer set solvers. The abstract framework simplifies the analysis of the correctness of algorithms and allows us to study the relationship between various algorithms by analyzing the differences in strategies of choosing a path in the graph. For example, the description of the SMODELS$_{cc}$ and SUP algorithms in this framework reflects their differences in a simple manner via distinct assignments of priorities to edges of the graph that characterize these systems. Also we used this framework to describe two algorithms for computing *Decision* and *FirstUIP* backjump clauses for the implementation of conflict-driven backjumping and learning. This formalism provided the transparent means for specifying these algorithms. We believe that the development of this abstract framework powerful enough to describe advanced features of answer set solvers in a simple manner will promote the use of these sophisticated features in more solvers. This work helped us design the new solver SUP, and we hope that in the future it will suggest designs of other systems for computing answer sets.

## Acknowledgments

We are grateful to Marco Maratea for bringing to our attention the work by Nieuwenhuis et al. (2006), to Vladimir Lifschitz for the numerous discussions, to Martin Gebser, Michael Gelfond, and Mirosław Truszczyński for valuable ideas and comments, to anonymous referees for their suggestions. The author was supported by the National Science Foundation under Grant IIS-0712113.